\documentclass[10pt,twocolumn,letterpaper]{article}

 \usepackage[pagenumbers]{cvpr} % To force page numbers, e.g. for an arXiv version

\definecolor{cvprblue}{rgb}{0.21,0.49,0.74}
\definecolor{citegreen}{rgb}{0.1,0.7,0.1}
\definecolor{grayline}{rgb}{0.5,0.5,0.5}
\definecolor{greenline}{rgb}{0.254,0.492,0.133}
\usepackage[pagebackref,breaklinks,colorlinks,citecolor=citegreen]{hyperref}
\usepackage{caption}
\usepackage{pifont}
\usepackage{multirow}
\usepackage{colortbl}
\usepackage{tabularx}
\usepackage{array} 
\usepackage{graphicx}

\makeatletter
\def\blfootnote{\xdef\@thefnmark{*}\@footnotetext}
\makeatother

\definecolor{best}{rgb}{1, 0.5, 0}
\definecolor{second}{rgb}{1, 1, 0}

\usepackage[capitalize]{cleveref}
\crefname{section}{Sec.}{Secs.}
\Crefname{section}{Section}{Sections}
\Crefname{table}{Table}{Tables}
\crefname{table}{Tab.}{Tabs.}

\def\paperID{8625} % *** Enter the Paper ID here
\def\confName{CVPR}
\def\confYear{2026}

\title{SwiftVGGT: A Scalable Visual Geometry Grounded Transformer \\for Large-Scale Scenes}

\author{{Jungho Lee \quad Minhyeok Lee \quad Sunghun Yang \quad Minseok Kang \quad Sangyoun Lee\vspace{+2mm}}\\
	{School of Electrical and Electronic Engineering, Yonsei University}\\
}

\begin{document}
\twocolumn[{
	\renewcommand\twocolumn[1][]{#1}
	\maketitle
	\begin{center}
		\centering
		\captionsetup{type=figure}
		\vspace{-6mm}
		\includegraphics[width=1\linewidth]{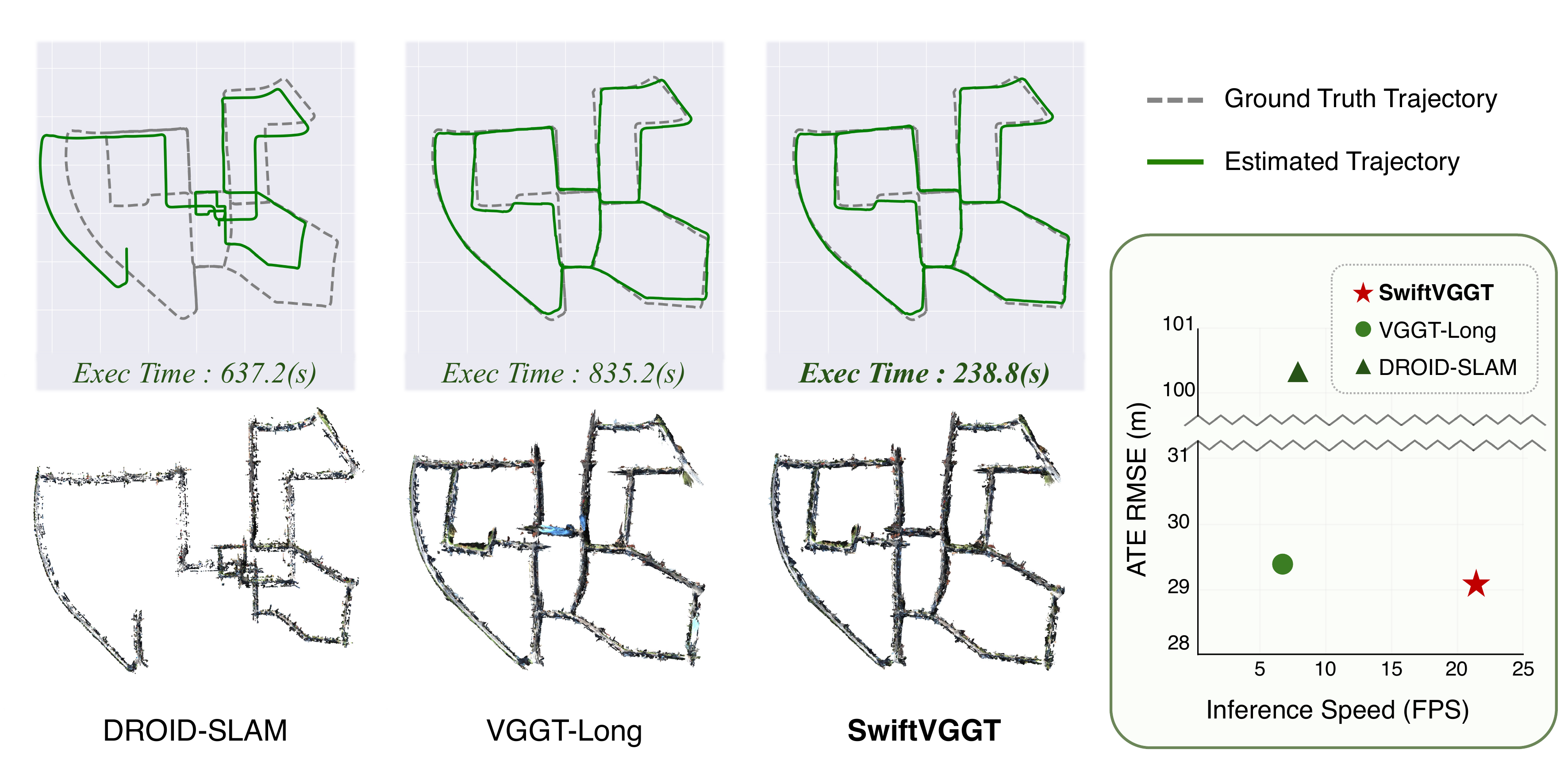}
		\vspace{-6mm}
		\caption{We propose SwiftVGGT, a method that achieves both high reconstruction quality and fast runtime. The scene shown is the KITTI dataset~\cite{kitti} sequence 00, consisting of 4,542 images. SwiftVGGT provides significantly faster processing while improving camera tracking accuracy and dense 3D reconstruction quality compared to prior approaches.}
		\label{fig:1_teaser}
	\end{center}
}]
\maketitle
\begin{abstract}
\blfootnote{Project Page: \href{https://Jho-Yonsei.github.io/SwiftVGGT/}{\textcolor{cvprblue}{\textit{https://Jho-Yonsei.github.io/SwiftVGGT/}}}}3D reconstruction in large-scale scenes is a fundamental task in 3D perception, but the inherent trade-off between accuracy and computational efficiency remains a significant challenge. Existing methods either prioritize speed and produce low-quality results, or achieve high-quality reconstruction at the cost of slow inference times. In this paper, we propose \textbf{SwiftVGGT}, a training-free method that significantly reduce inference time while preserving high-quality dense 3D reconstruction. To maintain global consistency in large-scale scenes, SwiftVGGT performs loop closure without relying on the external Visual Place Recognition (VPR) model. This removes redundant computation and enables accurate reconstruction over kilometer-scale environments. Furthermore, we propose a simple yet effective point sampling method to align neighboring chunks using a single Sim(3)-based Singular Value Decomposition (SVD) step. This eliminates the need for the Iteratively Reweighted Least Squares (IRLS) optimization commonly used in prior work, leading to substantial speed-ups. We evaluate SwiftVGGT on multiple datasets and show that it achieves state-of-the-art reconstruction quality while requiring only 33\% of the inference time of recent VGGT-based large-scale reconstruction approaches.
\end{abstract}    
\section{Introduction} \label{sec:1_introduction}

Despite recent advancements in 3D computer vision, predicting 3D geometry from videos captured in kilometer-scale environments, such as those found in autonomous driving scenarios, remains highly challenging. Modern large-scale Simultaneous Localization and Mapping (SLAM) systems~\cite{dpvo,orbslam,orbslam2,orbslam3} typically rely on feature-based sparse map representations to ensure real-time performance and computational efficiency. However, dense 3D reconstruction is essential for a deeper understanding of scene geometry, reliable obstacle avoidance, and safe motion planning in real-world autonomous operation. Although several recent works~\cite{droidslam,dpv-slam,gigaslam,dvso,dfvo} have attempted to address large-scale dense reconstruction, many of them still face practical limitations. They often require precise camera intrinsics, depend on complex multi-stage pipelines, or ultimately produce only semi-dense or sparse 3D maps due to computational bottlenecks. These constraints highlight the need for a method that can efficiently reconstruct dense 3D geometry at scale, without relying on heavy auxiliary modules or sacrificing inference speed.

Large-scale dense 3D reconstruction for fully autonomous driving environments requires a representation that can simultaneously handle \textit{long-range camera trajectory}, \textit{accumulated drift}, \textit{memory efficiency}, and \textit{fast inference speed}. Recent methods have attempted to leverage 3D vision foundation models~\cite{dust3r,mast3r,vggt} to ease this burden, but they typically address only a subset of these requirements and fail to cover all of them. For example, CUT3R~\cite{cut3r} and Fast3R~\cite{fast3r}, which build upon the transformer-based 3D foundation model DUSt3R~\cite{dust3r}, still suffer from memory capacity limitations when processing large-scale scenes. MASt3R-SLAM~\cite{mast3r-slam} improves reconstruction quality by incorporating MASt3R~\cite{mast3r} as a geometric prior, yet it often experiences tracking failures as the scene grows, leading to degraded trajectory estimation. More recent approaches have begun to adopt VGGT~\cite{vggt}, a foundation model that has shown strong performance in dense 3D perception tasks. However, VGGT is constrained by its intrinsic memory footprint, limiting the number of input images that can be processed simultaneously. FastVGGT~\cite{fastvggt} attempts to mitigate this issue, but it still encounters memory scalability problems when applied to kilometer-scale autonomous driving scenes.

Especially, VGGT-Long~\cite{vggtlong} addresses memory efficiency and drift mitigation by adopting a chunk-based Sim(3) alignment strategy, and handles loop closure in large-scale environments using a Visual Place Recognition (VPR) model~\cite{salad}. However, this approach still falls short of real-time performance, and we attribute this limitation to two main factors. First, chunk-to-chunk alignment relies on Iteratively Reweighted Least Squares (IRLS)~\cite{irls}-based Sim(3) optimization. Since each chunk contains a large number of 3D points, this repetitive optimization step becomes computationally expensive as shown in~\cref{tab:1_time_comparison}. Second, VGGT-Long detects loop closure by running a DINO~\cite{dinov2}-based VPR encoder~\cite{salad} that operates independently from the DINO encoder within VGGT itself, introducing additional computational overhead. These bottlenecks ultimately hinder the applicability of VGGT-Long in real-time autonomous driving scenarios, where both speed and scalability are essential.

In this paper, we introduce \textbf{SwiftVGGT}, a training-free approach that significantly accelerates inference for large-scale dense reconstruction without incurring additional memory cost, while maintaining or even improving reconstruction quality. Our method consists of two key components. First, we replace the iterative IRLS-based Sim(3) alignment with a single-step Sim(3)-based Singular Value Decomposition (SVD) procedure~\cite{umeyama}. To this end, we first align VGGT-predicted depth maps to a reference intrinsic scale, and then sample reliable points from the overlapping regions of neighboring chunks based on their aligned depth differences, enabling robust Sim(3) estimation. This sampling strategy ensures stable alignment while removing the need for repeated optimization iterations. Second, we remove the redundant computation introduced by the external VPR model by performing loop-closure detection directly with VGGT's DINO patch tokens. However, these features are not inherently suitable for place recognition, and naive usage fails to produce reliable loop correspondences. To address this, we introduce a feature transformation strategy that enables the tokens to serve effectively as a substitute for a dedicated VPR encoder. Together, these components enable SwiftVGGT to achieve state-of-the-art dense reconstruction and camera tracking performance, while delivering at least a 3$\times$ speedup in inference time compared to existing approaches.

\begin{table}[!t]
	\centering
	\caption{Runtime comparison for chunk alignment and loop detection on KITTI dataset~\cite{kitti} sequence 00. Our method is substantially faster by removing IRLS and the external VPR model.}
	\vspace{-2mm}
	\resizebox{0.9\columnwidth}{!}{
		\renewcommand{\arraystretch}{1.0}
		\scriptsize 
		\begin{tabular}{l||c|c}
			\toprule
			\multirow{2}{*}{Methods} & \multicolumn{2}{c}{~Elapsed Time (s)~} \\ \cmidrule{2-3}
			& Chunk Alignment & Loop Detection \\ \midrule 
			VGGT-Long  & 293.14 & 52.68 \\ [1.5pt]
			Ours	 & \textbf{23.18} (91.78\%$\downarrow$) & \textbf{1.19} (97.74\%$\downarrow$) \\ \bottomrule
			
		\end{tabular}
	}
	\vspace{-3mm}
	\label{tab:1_time_comparison}
\end{table}

To demonstrate the effectiveness of our model, we conduct experiments on the benchmarks: KITTI dataset~\cite{kitti}, Waymo Open dataset~\cite{waymo}, Virtual KITTI dataset~\cite{vkitti}. Our contributions can be summarized as follows:
\begin{itemize}
	\item [$\bullet$] We introduce a reliability-guided point sampling strategy that enables non-iterative Sim(3) alignment, replacing IRLS and significantly reducing alignment cost.
	\vspace{1mm}
	\item [$\bullet$] We propose a training-free loop closure mechanism that leverages VGGT encoder features, eliminating the external VPR module and redundant computation.
	\vspace{1mm}
	\item [$\bullet$] We demonstrate that SwiftVGGT achieves state-of-the-art large-scale dense reconstruction and camera tracking performance while achieving over 3$\times$ faster inference without additional memory overhead.
\end{itemize}
\section{Related Work} \label{sec:2_related_work}

\subsection{Learning-Based 3D Reconstruction} \label{sec:2.1_learning_based_3d_recon}

Building upon the foundations of traditional 3D reconstruction~\cite{orbslam,orbslam2,orbslam3,codeslam,gradslam,droidslam}, recent end-to-end learning-based approaches leverage transformer architectures to encode 3D scene priors directly from large-scale datasets. DUSt3R~\cite{dust3r} reconstructs view-consistent point maps without relying on camera calibration, while MASt3R~\cite{mast3r} enhances localization and Structure-from-Motion (SfM) performance by predicting additional per-pixel features that improve dense correspondence. CUT3R~\cite{cut3r} proposes a continuous 3D perception framework suitable for online, streaming input settings. Fast3R~\cite{fast3r} extends DUSt3R to process image collections exceeding 1,000 frames in a single forward pass, enabling large-scale scene reconstruction. More recently, VGGT~\cite{vggt} introduced a unified framework that jointly predicts camera intrinsics and extrinsics, depth maps, dense 3D point clouds from multiple input images, marking a new paradigm in feed-forward 3D vision. However, due to their inherent memory footprint, all the aforementioned methods are limited in the number of images it can process simultaneously, making it ineffective for kilometer-scale large-scene reconstruction scenarios.

\subsection{Large-Scale 3D Reconstruction} \label{sec:2.2_large_scale_3d_recon}

Traditional approaches to large-scale 3D reconstruction~\cite{orbslam,gigaslam,ldso,dpv-slam,droidslam} have primarily relied on SLAM-based pipelines. ORB-SLAM~\cite{orbslam} introduces a feature-based framework with keyframe selection, local and global bundle adjustment, and loop closure for large-scene mapping. DPV-SLAM~\cite{dpv-slam} models learned depth uncertainty and incorporates it into both tracking and mapping to achieve more consistent reconstruction. However, these methods generally produce sparse 3D maps. DROID-SLAM~\cite{droidslam} reconstructs comparatively denser maps, but it still faces limitations when scaling to kilometer-scale environments. More recently, VGGT-Long~\cite{vggtlong} leverages the VGGT to perform dense large-scale reconstruction, addressing sparsity issues inherent in SLAM-based mapping. Nevertheless, VGGT-Long does not fully exploit the feed-forward efficiency of VGGT, and its pipeline remains constrained by additional loop closure computation and iterative Sim(3) optimization, which limits real-time scalability.

\begin{figure*}[t]
	\centering
	\includegraphics[width=\textwidth]{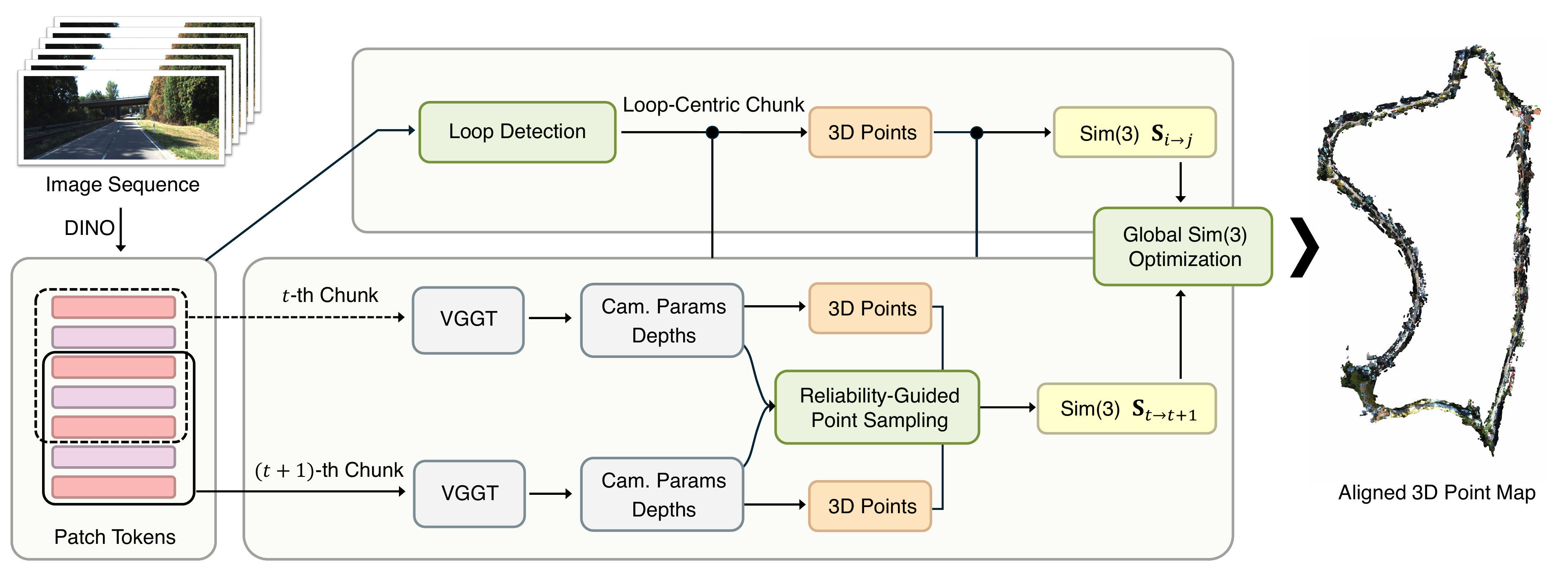}
	\vspace{-7mm}
	\caption{SwiftVGGT processes thousands of input images by dividing them into sliding-window chunks through VGGT~\cite{vggt}. To reduce inference time, we eliminate the IRLS optimization step by applying reliability-guided point sampling. Furthermore, we utilize the patch tokens obtained from the VGGT encoder~\cite{dinov2} for loop detection directly, which further decreases the overall inference cost.}
	\label{fig:2_pipeline}
	\vspace{-4mm}
\end{figure*}
\section{Method}
\label{sec:method}

\subsection{Overall Pipeline} \label{sec:3.1_overall_pipeline}

Our overall pipeline is based on VGGT-Long~\cite{vggtlong}, as shown in \cref{fig:2_pipeline}. Given a sequence of $N$ input images, we divide it into $T$ overlapping sliding-window chunks. We refer to each such chunk as a temporal chunk. Let the chunk size be $B$ and the overlap size be $O$. Then, the frame indices of the $t$-th chunk are defined as:$[(t-1)(B-O),\; (t-1)(B-O) + B].$ Each chunk is processed independently through VGGT, producing individual camera intrinsics/extrinsics, depth maps, and depth confidence maps.

For a pair of neighboring temporal chunks $\mathcal{C}_t$ and $\mathcal{C}_{t+1}$, we construct point maps by back-projecting depth maps using the estimated camera parameters. We then estimate a local Sim(3) transformation over the overlapping region of size $O$. To avoid the iterative IRLS optimization used in VGGT-Long, we introduce a \textit{reliability-guided point sampling} strategy that selects only stable points in the overlapping region of neighboring chunks, allowing a single-step Umeyama SVD~\cite{umeyama} for Sim(3) estimation (\cref{sec:3.2_point_sampling}). Moreover, since large-scale scenes consist of numerous temporal chunks, even small alignment errors can accumulate into noticeable drift. To address this, we propose a \textit{loop detection method that relies solely on VGGT's DINO patch tokens}, eliminating the external VPR model~\cite{salad} and avoiding redundant computation (\cref{sec:3.3_loop_detection}). As a result, removing IRLS iterations and VPR inference significantly reduces runtime, enabling fast reconstruction at kilometer-scale, as shown in~\cref{tab:1_time_comparison}. Finally, following VGGT-Long, we perform global Sim(3) optimization over all chunks and detected loops to ensure global consistency across the reconstructed scene (\cref{sec:3.4_global_optimization}).

\subsection{Reliability-Guided Point Sampling} \label{sec:3.2_point_sampling}

In this section, we describe a method that enables Sim(3) estimation using a single Umeyama step, removing the need for the iterative IRLS optimization used in VGGT-Long. VGGT outputs camera intrinsics/extrinsics, depth maps, and depth confidence maps for each frame. Since the estimated camera intrinsics vary across images, we first normalize all depth maps to a common reference intrinsic. This allows point maps from different chunks to be compared in a consistent scale. Given a source depth map $D_{\text{src}}$ and its source intrinsics $(f_{x,\text{src}}, f_{y,\text{src}})$, we rescale the depth to a reference intrinsic $(f_{x,\text{ref}}, f_{y,\text{ref}})$ as follows:
\begin{equation}
	D_{\text{reg}} = \frac{1}{2}\left(\frac{f_{x,\text{ref}}}{f_{x,\text{src}}} + \frac{f_{y,\text{ref}}}{f_{y,\text{src}}}\right) D_{\text{src}}.
\end{equation}
For convenience, we select the intrinsics of the first frame of the first temporal chunk $\mathcal{C}_{0}$ as the reference intrinsic. This intrinsic normalization ensures that the back-projected point clouds across chunks lie in the same metric scale, which is necessary before Sim(3) alignment.

Once the depth maps are normalized to the reference intrinsic, the overlapping regions of adjacent temporal chunks $\mathcal{C}_t$ and $\mathcal{C}_{t+1}$ exhibit consistent depth estimates. However, discrepancies remain in regions with very large depth, such as sky areas, as well as along object boundaries where sharp depth discontinuities occur. We mask out these regions and retain only the pixels whose depth differences are sufficiently small, treating their corresponding 3D points as reliable candidates for Sim(3) estimation. In addition, we leverage the depth confidence scores predicted to further filter out unstable regions. The final sampling mask used to select reliable points is defined as:
\begin{equation}
	\begin{aligned}
		\text{mask} = &
		\Big( |D_{\text{reg},t} - D_{\text{reg},t+1}| < \lambda_D \Big) \\
		&\land \Big( \gamma_t > \lambda_\gamma\,\mu_{\gamma_t} \Big)
		\land \Big( \gamma_{t+1} > \lambda_\gamma\,\mu_{\gamma_{t+1}} \Big),
	\end{aligned}
\end{equation}
where $D_{\text{reg},t}$ is the depth map of chunk $\mathcal{C}_t$ rescaled to the reference intrinsic, $\lambda_D$ and $\lambda_\gamma$ are the depth-difference and confidence thresholds; $\gamma_t$ is the depth confidence, and $\mu_{\gamma_t}$ its mean. The symbol $\land$ denotes the logical \textsc{and} operator.

Since the overlapping region already provides a reliable correspondence between the two chunks $\mathcal{C}_{t}$ and $\mathcal{C}_{t+1}$, we assume these selected points are well-aligned and directly apply the Umeyama algorithm to estimate the Sim(3) transformation $\mathbf{S}_{t\rightarrow t+1}$, without iterative IRLS refinement. This removes the IRLS optimization overhead and significantly reduces runtime, as shown in \cref{tab:1_time_comparison}.

\subsection{Loop Detection from VGGT Encoder Features} \label{sec:3.3_loop_detection}

In this section, we describe our loop detection algorithm, which operates without any external VPR encoder. Prior large-scale reconstruction systems~\cite{vggtlong} typically employ a dedicated VPR encoder~\cite{salad} and an additional encoder~\cite{vggt} for 3D reconstruction, resulting in redundant computation and increased inference time. In contrast, we leverage VGGT's DINO~\cite{dinov2} patch tokens and transform them into global descriptors suitable for loop detection. Although these tokens are not originally designed for place recognition, their strong semantic expressiveness inherited from DINO and the multi-view consistency reinforced through VGGT’s frame and global attention make them well suited for robust loop correspondence estimation.

Unlike prior VPR works~\cite{netvlad,cosplace,transvpr,mixvpr} that design dedicated architectures and training objectives specifically for place recognition, our goal is not to improve standalone VPR benchmarks, but to repurpose VGGT's DINO transformer for loop-closure detection without any additional training. To the best of our knowledge, we are the first to demonstrate that the encoder of a 3D vision foundation model can be transformed into a VPR-style loop descriptor purely through classical feature normalization (signed power normalization and PCA whitening), without modifying or fine-tuning the backbone. While RootSIFT~\cite{rootsift}-style normalization and PCA whitening are well-established in image retrieval and VPR, prior works typically apply them on descriptors explicitly trained for place recognition. %In contrast, we show that the same tools, when applied to VGGT's 3D-centric patch tokens, are sufficient to obtain robust loop-closure constraints for large-scale 3D reconstruction.

Patch tokens $X_i$ for each image $I_i$ are defined as follows:
\begin{equation} \label{eq:1_patch_token}
%	\vspace{-1mm}
	X_i \in \mathbb{R}^{K \times d}, \quad X_i = [x_{i,1}, x_{i,2}, \cdots, x_{i,K}]^{\top},
\end{equation}
where $K$ denotes the number of tokens and $d$ the token feature dimension. We first apply $\ell_2$-normalization along the token dimension and then average across tokens to obtain an initial global representation $g_i^{(0)} \in \mathbb{R}^{d}$.

To prevent a few dominant features from overwhelming the representation, we apply signed power normalization~\cite{signedpower}, which smooths the feature distribution while preserving sign information. This is conceptually similar to the normalization used in RootSIFT~\cite{rootsift}, and helps produce more balanced and discriminative descriptors:
\begin{equation} \label{eq:2_singed_power}
	g_i^{(1)} = \text{sign}(g_i^{(0)}) \cdot |g_i^{(0)}|^{\beta}, \quad \hat{g}_i^{(1)} = \frac{g_i^{(1)}}{\|g_i^{(1)}\|_2},
\end{equation}
where we set $\beta = 0.5$ to suppress overly large responses while relatively amplifying smaller ones, which helps distribute feature magnitudes more evenly and broadens the similarity range among loop candidates.

Since VGGT's DINO transformer is optimized for 3D reconstruction, its global features capture coarse scene priors and dataset biases. This causes unrelated frames to share similarly shaped global descriptors, which raises cosine similarity and produces hubness in the feature space. To mitigate this issue, we reduce the dominant shared directions in the descriptor space. Let $G$ denote the matrix whose rows are the descriptors $g_i^{(1)}$. We perform eigen-decomposition on $G$ and remove the top $r$ principal components associated with the largest eigenvalues, retaining only $d'$ output dimensions. Let $Q$ and $\Lambda$ denote the eigenvectors and eigenvalues after discarding these $r$ dominant components. We then construct the whitening matrix $W = Q \Lambda^{-\frac{1}{2}} \in \mathbb{R}^{d \times d'}.$ This transform rescales the remaining subspace and yields a more discriminative representation. The final global descriptor $z_i$ is then given by:
\begin{equation} \label{eq:3_global_descriptor}
	z_i = (g_i^{(1)} - \mu) W, \quad \hat{z}_i = \frac{z_i}{\|z_i\|_2} \in \mathbb{R}^{d'},
\end{equation}
where $\mu$ denotes the mean descriptor of the scene.

Following VGGT-Long~\cite{vggtlong}, loop detection is performed via cosine similarity between normalized descriptors. We stack the $N$ descriptors into $Z \in \mathbb{R}^{N \times d'}$ and derive loop candidates from the similarity matrix $ZZ^\top$, retaining only pairs above a similarity threshold and separated in time. Finally, frame pairs whose cosine similarity exceeds a threshold and whose frame indices are sufficiently separated are regarded as loop candidates. Non-Maximum Suppression (NMS) is then applied to prevent overlapping detections.

For each detected loop pair $(I_i, I_j)$, we construct a loop-centric batch by concatenating the image sequences centered around indices $i$ and $j$, which preserves spatial correspondence while not being temporally continuous. This batch is passed through VGGT to produce a loop-centric chunk $\mathcal{C}_{\text{loop}}$, which is then aligned with the corresponding temporal chunks $\mathcal{C}_{[i]}$ and $\mathcal{C}_{[j]}$, where $\mathcal{C}_{[i]}$ and $\mathcal{C}_{[j]}$ denote the chunks that contain frames $I_i$ and $I_j$, respectively. Finally, we obtain the loop-closing Sim(3) transformation using the transformations from the temporal chunks $\mathcal{C}_{[i]}$ and $\mathcal{C}_{[j]}$ to the loop-centric chunk $\mathcal{C}_{\text{loop}}$. Let $\mathbf{S}_{i,\text{loop}}$ and $\mathbf{S}_{j,\text{loop}}$ denote the Sim(3) transformations that align $\mathcal{C}_{[i]}$ and $\mathcal{C}_{[j]}$ to $\mathcal{C}_{\text{loop}}$, respectively. The resulting loop-closing transformation is then computed as:
\begin{equation}
	\mathbf{S}_{i\rightarrow j} = \mathbf{S}_{j,\text{loop}} \, \mathbf{S}_{i,\text{loop}}^{-1}.
\end{equation}
This yields a direct spatial constraint between temporally distant chunks that contain loop-consistent observations.

\subsection{Global Sim(3) Optimization} \label{sec:3.4_global_optimization}

Following previous approaches~\cite{gigaslam,dpv-slam,scaledrift}, we perform a global Sim(3) optimization over all temporal chunks to ensure geometric consistency across both sequential and loop-closing constraints. Let $\{\mathbf{S}_t\}_{t=1}^{T}$ denote the Sim(3) transformations associated with the $T$ temporal chunks. The relative Sim(3) transformation obtained between neighboring temporal chunks is denoted by $\mathbf{S}_{t\rightarrow t+1}$ (\cref{sec:3.2_point_sampling}), and the loop-closing constraint between non-adjacent chunks is represented by $\mathbf{S}_{i\rightarrow j}$ (\cref{sec:3.3_loop_detection}).

To solve for $\{\mathbf{S}_t\}$, we map each Sim(3) element to its Lie algebra using the $\log_{\text{Sim(3)}}$, producing an unconstrained 7D residual. We then apply Levenberg--Marquardt (LM)~\cite{levenberg1944,marquardt1963} to minimize a nonlinear least-squares objective:
\begin{equation}
	\vspace{-1mm}
	\begin{aligned}
			\min_{\{\mathbf{S}_t\}}\mathcal{L}(\{\mathbf{S}_t\}) =  \;&
			\overbrace{\sum_{t \in T}
					\left\| \log\left( \mathbf{S}_{t\rightarrow t+1}^{-1} \mathbf{S}_t^{-1} \mathbf{S}_{t+1} \right) \right\|_{2}^{2}}^{\text{Constraints for Temporal Chunks}} ~~+ \\%[4pt]
			& \underbrace{\sum_{(i,j)\in \mathcal{E}_{\text{loop}}}
					\left\| \log\left( \mathbf{S}_{i\rightarrow j}^{-1} \mathbf{S}_i^{-1} \mathbf{S}_j \right) \right\|_{2}^{2}}_{\text{Constraints for Loop-Centric Chunks}},
		\end{aligned}
	\vspace{-1mm}
\end{equation}
where the first term enforces consistency between adjacent temporal chunks, while the second term incorporates long-range constraints derived from detected loop closures, where $\mathcal{E}_{\text{loop}}$ denotes the set of loop edges. This joint optimization effectively distributes accumulated drift across the entire sequence and ensures global coherence of the reconstructed 3D geometry.

\begin{table*}[!t] 
	\begin{center}
		\caption{Performance comparison on the KITTI Odometry dataset~\cite{kitti}. The $^{\dagger}$ denotes sequences containing loops, and \textit{Dense}$^\diamondsuit$ indicates semi-dense 3D reconstruction. The \colorbox{best!25}{orange} and \colorbox{second!35}{yellow} cells respectively indicate the highest and second-highest value.}
		\vspace{-3mm}
		\resizebox{\linewidth}{!}{
			\centering
			\setlength{\tabcolsep}{5pt}
			\scriptsize
			\begin{tabular}{l||c|c|ccccccccccc|cc}
				\toprule 
				Methods				& Calib.		& Recon.		& 00$^\dagger$		& 01		& 02$^\dagger$		& 03		& 04		& 05$^\dagger$		& 06$^\dagger$		& 07$^\dagger$		& 08$^\dagger$		& 09$^\dagger$		& 10		& Average 		& FPS\\ \midrule \midrule
				DPV-SLAM~\cite{dpv-slam}		& \ding{51}		& \textit{Sparse}		& 112.80		& 11.50		& 123.53		& 2.50		& 0.81		& 57.8		& 54.86		& 18.77		& 110.49		& 76.66		& 13.65		& 53.03 & 31.37 \\ \midrule
				MASt3R-SLAM~\cite{mast3r-slam}		& \ding{55}		& \textit{Dense}		& \multicolumn{11}{c|}{\textit{Tracking Lost}}		& -- & -- \\ [1pt]
				CUT3R~\cite{cut3r}		& \ding{55}		& \textit{Dense}		& \multicolumn{11}{c|}{\textit{Out of Memory}}		& -- & --\\ [1pt]
				Fast3R~\cite{fast3r}		& \ding{55}		& \textit{Dense}		& \multicolumn{11}{c|}{\textit{Out of Memory}}		& -- & --\\ [1pt]
				FastVGGT~\cite{fastvggt}		& \ding{55}		& \textit{Dense}		& \multicolumn{11}{c|}{\textit{Out of Memory}}		& -- & --\\ \midrule
				DROID-SLAM~\cite{droidslam}		& \ding{51}		& \textit{Dense}$^\diamondsuit$		& 92.10		& 344.6		& 107.61		& \cellcolor{best!25}2.38		& \cellcolor{best!25}1.00		& 118.5		& 62.47		& 21.78		& 161.60		& 72.32		& 118.70		& 100.28 & \cellcolor{second!35}8.08 \\ [1pt]
				VGGT-Long~\cite{vggtlong}		& \ding{55}		& \textit{Dense}		& \cellcolor{second!35}9.87		& \cellcolor{second!35}111.06		& \cellcolor{second!35}37.56		& \cellcolor{second!35}4.89		& \cellcolor{second!35}3.75		& \cellcolor{best!25}{9.09}		& \cellcolor{best!25}{7.47}		& \cellcolor{best!25}{4.02}		& \cellcolor{best!25}{62.86}		& \cellcolor{second!35}47.48		& \cellcolor{best!25}{25.49}		& \cellcolor{second!35}29.41 & 6.91 \\ \midrule
				Ours					& \ding{55}		& \textit{Dense}		& \cellcolor{best!25}{8.17}		& \cellcolor{best!25}102.53		& \cellcolor{best!25}{36.49}		& 8.12		& 4.88		& \cellcolor{second!35}11.94		& \cellcolor{second!35}8.88		& \cellcolor{second!35}5.01		& \cellcolor{second!35}64.68		& \cellcolor{best!25}{44.13}		& \cellcolor{second!35}26.18		& \cellcolor{best!25}{29.18} & \cellcolor{best!25}20.73 \\ \bottomrule
			\end{tabular}
		}
		\label{tab:2_comparison_kitti_performance}
	\end{center}
	\vspace{-4mm}
\end{table*}

\begin{table*}[!t] 
	\begin{center}
		\caption{Performance comparison on the Waymo Open dataset~\cite{waymo}.}
		\vspace{-3mm}
		\resizebox{\linewidth}{!}{
			\centering
			\setlength{\tabcolsep}{3pt}
			\scriptsize
			\begin{tabular}{l||c|c|ccccccccc|cc}
				\toprule 
				Methods				& Calib.		& Recon.		& 163453191		& 183829460		& 315615587		& 346181117		& 371159869		& 405841035		& 460417311		& 520018670		& 610454533		& Average 		& FPS \\ \midrule \midrule
				Fast3R~\cite{fast3r}		& \ding{55}		& Dense		& \multicolumn{9}{c|}{\textit{Out of Memory}}		& -- & --\\ \midrule
				DROID-SLAM~\cite{droidslam}		& \ding{51}		& Dense$^\diamondsuit$		& 3.705		& \cellcolor{best!25}0.301		& \cellcolor{best!25}0.447		& 8.653		& 9.320		& 7.621		& 4.170		& \textit{T.L.}		& \cellcolor{best!25}0.264		& 4.310		& 7.21	 \\ [1pt]
				MASt3R-SLAM~\cite{mast3r-slam}		& \ding{55} 		& Dense		& 4.500		& \cellcolor{second!35}0.556		& \cellcolor{second!35}1.833		& 12.544		& 8.601		& \cellcolor{best!25}1.412		& 5.428		& 7.910		& \cellcolor{second!35}1.195		& 4.887		& 5.47	\\ [1pt]
				CUT3R~\cite{cut3r}		& \ding{55}		& Dense		& 8.781		& 3.810		& 5.790		& 24.015		& 13.070		& 7.261		& 13.206		& 8.597		& 3.229		& 9.751		& 8.38	\\ [1pt]
				FastVGGT~\cite{fastvggt}		& \ding{55}		& Dense		& 8.380		& 2.639		& 3.430		& \cellcolor{second!35}2.965		& 9.915		& 3.611		& 4.643		& 18.45		& 2.414		& 6.272		& \cellcolor{best!25}11.38	\\ [1pt]
				VGGT-Long~\cite{vggtlong}		& \ding{55}		& Dense		& \cellcolor{best!25}3.086		& 2.544		& 2.336		& 4.007		& \cellcolor{best!25}4.045		& 3.141		& \cellcolor{second!35}2.867		& \cellcolor{second!35}3.407		& 2.333		& \cellcolor{second!35}3.085		& 1.97 \\ \midrule
				Ours					& \ding{55}		& Dense		& \cellcolor{second!35}3.106		& 2.447		& 2.346		& \cellcolor{best!25}2.719		& \cellcolor{second!35}4.080		& \cellcolor{second!35}3.106		&  \cellcolor{best!25}2.841		& \cellcolor{best!25}2.832		& 2.210		& \cellcolor{best!25}2.854		& \cellcolor{second!35}8.41	 \\ \bottomrule
			\end{tabular}
		}
		\label{tab:4_comparison_waymo_performance}
	\end{center}
	\vspace{-5mm}
\end{table*}

\section{Experiments}
\label{sec:experiments}

\subsection{Datasets and Evaluation Metrics}\label{sec:metrics}

\paragraph{Dataset.}SwiftVGGT is evaluated on three datasets: the KITTI Odometry dataset~\cite{kitti}, the Waymo Open dataset~\cite{waymo}, and the Virtual KITTI dataset~\cite{vkitti}. The KITTI dataset consists of 22 kilometer-scale road scenes, among which scenes 00–10 provide ground-truth camera poses, while scenes 11–21 do not. The Waymo Open dataset contains non-loop sequences of road scenes, each approximately 300 meters in length. The Virtual KITTI dataset is a synthetic urban environment generated from seed images of the real KITTI dataset, consisting of five distinct scenes, each under six different environmental conditions.

\vspace{-2mm}
\paragraph{Metrics.}For all datasets, we evaluate the camera tracking performance using the Absolute Trajectory Error (ATE) RMSE measured in meters. The performance on the Virtual KITTI dataset is presented in our \textbf{Appendix}. Since ground-truth poses are unavailable for KITTI scenes 11–21, we generate pseudo ground truth using the LiDAR-based SLAM method, PIN-SLAM~\cite{pinslam}, and evaluate our method on the loop-containing sequences to verify the effectiveness of our VPR-free loop detection. For the Waymo Open dataset, we follow VGGT~\cite{vggt} and measure three metrics: Accuracy (the minimum Euclidean distance from prediction to ground truth), Completeness (the inverse measure), and their mean, the Chamfer Distance (see \textbf{Appendix}). For KITTI and Virtual KITTI datasets, where point-level ground truth is unavailable and LiDAR scan density and range are limited, qualitative evaluation is conducted instead.

\subsection{Implementation Details}

SwiftVGGT is built upon the VGGT~\cite{vggt} backbone and operates in a fully training-free manner without any additional modules. Since we directly utilize VGGT’s depth outputs for point sampling, the point head is omitted; instead, the depth head alone is used to generate 3D points from the predicted depth and camera parameters. Following the same temporal configuration as VGGT-Long~\cite{vggtlong}, we set the temporal chunk size $B$ to 75 for the KITTI~\cite{kitti} dataset and 60 for the Waymo Open~\cite{waymo} dataset, with an overlap size $O$ of 30. For reliability-guided point sampling, we set the depth-difference threshold $\lambda_D = 0.2$ and the depth-confidence threshold $\lambda_{\gamma} = 0.5$. To perform loop detection using only VGGT's DINO transformer, we follow the PCA-whitening strategy. Specifically, we remove the top $r=1$ principal component from the descriptor matrix $G$ and retain the feature dimension $d' = 512$. The chunk size for loop-centric processing is set to 40. 

All experiments are conducted on a workstation equipped with an AMD Ryzen Threadripper 7960X CPU, 128 GB of RAM, and a single NVIDIA RTX 4090 GPU.

\begin{table}[!t] 
	\begin{center}
		\caption{Performance comparison on the KITTI Odometry dataset~\cite{kitti}, including sequences that contain loops, using pseudo ground truth camera pose generated by PIN-SLAM~\cite{pinslam}.}
		\vspace{-2mm}
		\resizebox{\columnwidth}{!}{
			\centering
			\setlength{\tabcolsep}{3pt}
			\scriptsize
			\begin{tabular}{l||ccccc|c}
				\toprule 
				Methods				& 13		& 15		& 16		& 18		& 19		& Average 	\\ \midrule \midrule
				DPV-SLAM~\cite{dpv-slam}				& 48.40		& 39.97		& 30.41		& 11.58		& 182.82				& 62.64 \\ \midrule
				MASt3R-SLAM~\cite{mast3r-slam}				& \multicolumn{5}{c|}{\textit{Tracking Lost}}		& --  \\
				CUT3R~\cite{cut3r}			& \multicolumn{5}{c|}{\textit{Out of Memory}}		& --\\
				Fast3R~\cite{fast3r}		& \multicolumn{5}{c|}{\textit{Out of Memory}}		& -- \\
				FastVGGT~\cite{fastvggt}			& \multicolumn{5}{c|}{\textit{Out of Memory}}		& -- \\ \midrule
				DROID-SLAM~\cite{droidslam}		& 53.73		& 51.20		& 34.15		& 17.96		& 229.36		& 77.28  \\
				VGGT-Long~\cite{vggtlong}		& \cellcolor{second!35}7.76		& \cellcolor{second!35}6.33		& \cellcolor{second!35}18.71		& \cellcolor{best!25}6.84		& \cellcolor{second!35}118.10			& \cellcolor{second!35}31.55  \\ \midrule
				Ours				& \cellcolor{best!25}6.72		& \cellcolor{best!25}6.30		& \cellcolor{best!25}16.27		& \cellcolor{second!35}7.60		& \cellcolor{best!25}111.73		& \cellcolor{best!25}{29.72} \\ \bottomrule
			\end{tabular}
		}
		\label{tab:3_comparison_kitti_loop_performance}
	\end{center}
	\vspace{-7mm}
\end{table}

\begin{figure*}[t]
	\centering
	\includegraphics[width=0.9\textwidth]{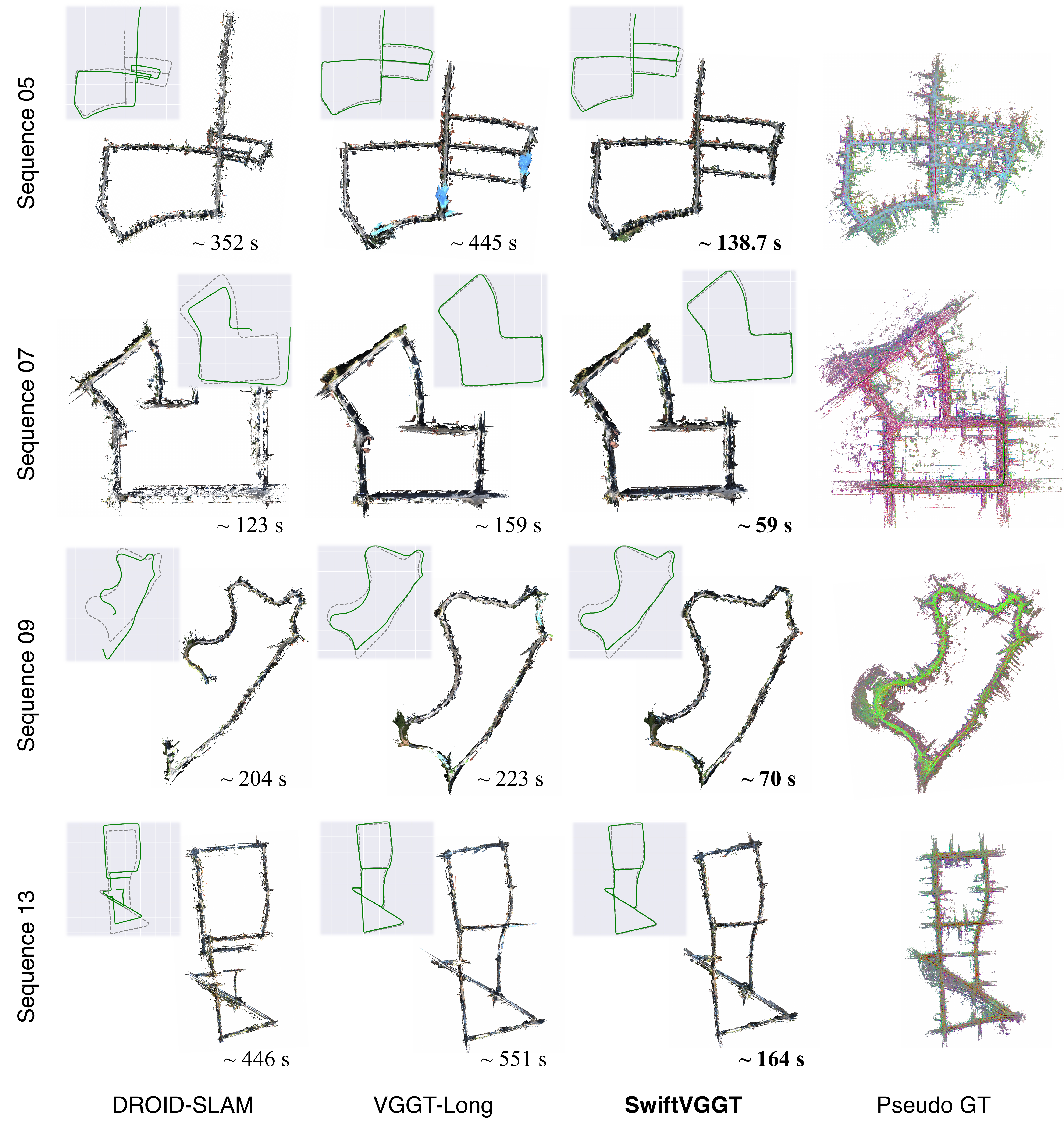}
	\vspace{-3mm}
	\caption{Comparison of dense 3D reconstruction on KITTI scenes containing loops. The pseudo ground-truth point cloud is obtained by combining the LiDAR point cloud with camera poses either provided by the dataset or estimated by PIN-SLAM~\cite{pinslam}. The \textcolor{grayline}{\textbf{gray}} dashed line represents the ground-truth trajectory, while the \textcolor{greenline}{\textbf{green}} solid line denotes the estimated trajectory.}
	\label{fig:4_comparison_kitti}
	\vspace{-3mm}
\end{figure*}

\subsection{Camera Tracking}

For quantitative evaluation, we compare the Absolute Trajectory Error (ATE) RMSE (m) on KITTI scenes 00–10 against other dense 3D reconstruction models. As shown in~\cref{tab:2_comparison_kitti_performance}, existing 3D vision foundation model-based methods such as MASt3R-SLAM~\cite{mast3r-slam}, CUT3R~\cite{cut3r}, Fast3R~\cite{fast3r}, and FastVGGT~\cite{fastvggt} often fail to process long sequences due to excessive memory usage, resulting in tracking loss or out-of-memory errors. DPV-SLAM~\cite{dpv-slam} employs DBoW2~\cite{dbow2}-based ORB features for loop detection, but demonstrates inferior performance on loop-containing sequences, as reported in~\cref{tab:2_comparison_kitti_performance}. Although DPV-SLAM runs relatively fast, it suffers from limited camera tracking accuracy and produces only sparse 3D reconstructions. DROID-SLAM~\cite{droidslam} achieves denser reconstructions than traditional feature-based SLAM systems, yet still performs worse and less densely than recent 3D vision foundation models. In contrast, our SwiftVGGT achieves consistently strong performance across all KITTI scenes, providing dense 3D reconstruction while running approximately three times faster than VGGT-Long~\cite{vggtlong}.

As shown in~\cref{tab:3_comparison_kitti_loop_performance}, we further evaluate on KITTI scenes 11–21, for which no ground-truth camera poses are provided. Following prior practice, we generate pseudo ground truth using the LiDAR-based SLAM method, PIN-SLAM~\cite{pinslam}. To validate the effectiveness of our VPR-free loop detection, we focus on sequences that contain loop closures. While the performance of DPV-SLAM and DROID-SLAM significantly deteriorates under these challenging conditions, both VGGT-Long and our methods maintain stable tracking accuracy. However, as reported in~\cref{tab:2_comparison_kitti_performance}, our method runs approximately three times faster while also achieving slightly better trajectory accuracy. This demonstrates that our loop detection strategy performs competitively with methods that rely on explicit VPR modules, and even surpasses traditional SLAM-based approaches.

As summarized in~\cref{tab:4_comparison_waymo_performance}, we also compare performance on the loop-free Waymo Open dataset~\cite{waymo}. Since this dataset consists of relatively short sequences, methods such as MASt3R-SLAM, CUT3R, and FastVGGT run without out-of-memory issues. FastVGGT achieves the fastest inference time among all methods, but its camera tracking accuracy is less than half of ours. Similarly, CUT3R demonstrates moderate speed but performs considerably worse in accuracy. While VGGT-Long achieves relatively strong tracking results, it remains approximately four times slower than our proposed SwiftVGGT.

\subsection{3D Reconstruction}

As discussed in the~\cref{sec:metrics}, the KITTI dataset lacks point-level ground truth, and its LiDAR scans are limited in both density and range. Therefore, we qualitatively evaluate the 3D reconstruction results of SwiftVGGT. We generate pseudo ground-truth point clouds for the entire scene using LiDAR point clouds combined with either the provided camera parameters or those estimated by PIN-SLAM~\cite{pinslam}. As shown in~\cref{fig:4_comparison_kitti}, although the pseudo ground truth and the reconstructed outputs exhibit differences in point density and distribution, the output geometry is sufficiently detailed for qualitative comparison. Note that DPV-SLAM~\cite{dpv-slam} produces overly sparse reconstructions and is therefore excluded from qualitative comparison. DROID-SLAM~\cite{droidslam} fails to consistently detect loop closures across evaluated scenes, while VGGT-Long~\cite{vggtlong} and our proposed SwiftVGGT successfully recover accurate loops and reconstruct geometries closely aligned with the pseudo ground truth. Moreover, SwiftVGGT achieves about 3$\times$ faster inference speed, demonstrating its superior practicality for large-scale 3D scene reconstruction. Additional qualitative comparisons on loop-free scenes are provided in the \textbf{Appendix}.

\begin{table}[!t] 
	\begin{center}
		\caption{Ablation study on chunk alignment. RGPS denotes the reliability-guided point sampling, and $\lambda_D$ represents the depth-difference threshold.}
		\vspace{-3mm}
		\resizebox{0.9\columnwidth}{!}{
			\centering
			\setlength{\tabcolsep}{5pt}
			\scriptsize
			\begin{tabular}{ccc|cc}
				\toprule 
				\multicolumn{2}{c}{Chunk Alignment}		& \multirow{2}{*}{$\lambda_D$}		& \multirow{2}{*}{ATE RMSE (m)}		& \multirow{2}{*}{Elapsed Time (s)} 	\\ [1pt]
				~~IRLS				& RGPS~									&																	&	&  \\ \midrule \midrule
				& 																	& --															&	32.95					& 111.4				\\ \midrule [1pt] 
				\ding{51} 		& 												& --															&	29.23					& 243.4				\\ [1pt]
				& \ding{51}												& 0.1															&	29.64					& \textbf{105.6}				\\ [1pt]
				& \ding{51}												& 0.2															&	\textbf{29.18}					& 105.8				\\ [1pt]
				& \ding{51}												& 0.3															&	29.95					& 107.2				\\ [1pt]
				& \ding{51} 											& 0.4															&	30.04					& 108.3 				\\ \bottomrule
			\end{tabular}
		}
		\label{tab:5_ablation_chunk_alignment}
	\end{center}
	\vspace{-7mm}
\end{table}

\subsection{Ablation Study}

\paragraph{Chunk Alignment.}To validate the effectiveness of our proposed chunk alignment strategy, we conduct an ablation study on the reliability-guided point sampling method using KITTI sequences 00--10, as summarized in~\cref{tab:5_ablation_chunk_alignment}. When Umeyama alignment is applied directly to all overlapping points between temporal chunks without IRLS or our point sampling, the ATE worsens, and the runtime increases slightly due to the maximal number of points involved. Incorporating IRLS improves ATE but more than doubles the inference time, making it impractical for large-scale settings. In contrast, our point sampling method not only yields better ATE than using all points but also achieves faster alignment due to the reduced and more reliable point set. We additionally sweep the depth-difference threshold~$\lambda_D$ and select $\lambda_D = 0.2$, which provides the best balance between accuracy and runtime.

\begin{table}[!t] 
	\begin{center}
		\caption{Ablation study on loop detection methods.}
		\vspace{-3mm}
		\resizebox{0.9\columnwidth}{!}{
			\centering
			\setlength{\tabcolsep}{5pt}
			\scriptsize
			\begin{tabular}{cc|cc}
				\toprule 
				\multicolumn{2}{c|}{Loop Detection}				& \multirow{2}{*}{ATE RMSE (m)}		& \multirow{2}{*}{Elapsed Time (s)} 	\\ [1pt]
				VPR~\cite{salad}				& Ours (\cref{sec:3.3_loop_detection})																			&	&  \\ \midrule \midrule
				\ding{51} 		& 															&	27.40					& 165.5				\\ [1pt]
				& \ding{51}																	&	\textbf{27.33}					& \textbf{131.7}				\\ \bottomrule
			\end{tabular}
		}
		\label{tab:6_ablation_loop_detection}
	\end{center}
	\vspace{-6mm}
\end{table}

\vspace{-3mm}
\paragraph{Loop Detection.}To evaluate the effectiveness of our loop detection method using only VGGT's DINO patch tokens (\cref{sec:3.3_loop_detection}), we perform the ablation study presented in~\cref{tab:6_ablation_loop_detection}. 
Since loop-closure performance can only be assessed on sequences that contain actual loops, we use KITTI Odometry sequences 00, 02, 05, 06, 07, 08, 09, 13, 15, 16, and 18. Both the VPR-based baseline and our method detect the same loop locations, resulting in nearly identical tracking performance. Minor discrepancies arise because different loop pairs may correspond to the same physical loop closure. Furthermore, our approach eliminates the need to run a separate VPR model, reducing the runtime by approximately 30 seconds on average for the evaluated sequences.
\section{Limitations}
\label{sec:limitations}

Although our method achieves both higher accuracy and faster runtime compared to existing SLAM-based approaches, it estimates camera parameters without bundle adjustment, as our primary goal is dense 3D reconstruction rather than precise pose refinement. Consequently, accumulated drift cannot be fully corrected, which may lead to distortions in camera trajectories and reconstructed geometry. Incorporating bundle adjustment into SwiftVGGT presents a promising direction for future work and may enable a new paradigm of large-scale structure-from-motion.
\section{Conclusion}
\label{sec:conclusion}

We propose SwiftVGGT, a fast and training-free large-scale 3D reconstruction framework that significantly accelerates feed-forward reconstruction pipelines while maintaining high-quality camera tracking and dense geometric accuracy. Our method provides two key contributions: (1) a reliability-guided point sampling strategy, and (2) an efficient loop detection mechanism. Extensive experiments demonstrate that SwiftVGGT achieves strong camera tracking performance, accurate large-scale 3D geometry, and robust loop recovery, all while running 3$\times$ faster than existing VGGT-based approaches. Qualitative and quantitative results confirm that SwiftVGGT provides a superior balance between speed and accuracy, making it highly practical for large-scale 3D perception.

\begin{center}
	\textbf{\Large{Appendix}}
	\vspace{5mm}
\end{center}
\appendix

\section{How Do VGGT's DINO Enable VPR?}

Although VGGT~\cite{vggt} was never designed for Visual Place Recognition (VPR), we find that its DINO~\cite{dinov2}-based encoder naturally contains several properties that make it unexpectedly effective for place-level matching once appropriately normalized. In this section, we analyze why this phenomenon occurs and explain how our lightweight post-processing turns VGGT’s reconstruction-centric tokens into competitive loop descriptors.
\vspace{-5mm}
\paragraph{VGGT's DINO Tokens.}The VGGT encoder is initialized from DINO, a self-supervised vision transformer known to produce highly semantic, spatially consistent patch tokens. These tokens capture mid-level structures such as building facades, vegetation patterns, and road geometry--exactly the type of cues that are robust to illumination changes and viewpoint variations in VPR. Whereas traditional VPR models~\cite{transvpr,cosplace,mixvpr,salad} explicitly learn such invariances through contrastive training, VGGT inherits them ``for free'' from DINO. This provides a strong initialization for transforming the patch tokens into global descriptors.
\vspace{-5mm}
\paragraph{VGGT's Frame/Global Attention.}Unlike standalone DINO, VGGT is trained for multi-view 3D reasoning using both frame attention and global attention. These mechanisms encourage \textit{(1) view-to-view feature consistency}, \textit{(2) stable representations for the same region observed under different poses}, \textit{(3) suppression of transient or dynamic details}. This implicit geometric alignment effect makes the encoder features more stable across large camera motions, which is valuable for long-range place recognition.
\vspace{-5mm}
\paragraph{Normalization.}Despite these strengths, raw VGGT tokens cannot be used directly for VPR because: (1) they contain strong low-frequency scene priors (dataset bias); (2) Dominant feature directions lead to hubness, where unrelated frames appear spuriously similar; (3) VGGT is optimized for 3D reconstruction, not discrimination across places. Hence, naive averaging of DINO tokens performs poorly as shown in~\cref{tab:additional_ablation_loop_detection}. A transformation is required to emphasize discriminative directions while suppressing dataset-shared ones.

Our transformation relies only on classical, training-free normalization techniques. First, signed power normalization step reduces the dominance of large-magnitude components and amplifies weaker ones. Although reminiscent of RootSIFT~\cite{rootsift}, classical VPR methods rarely apply it on transformer global descriptors. In our case, it significantly stabilizes the descriptor distributions across scenes. Second, while PCA whitening is common in retrieval, it is almost always applied together with a learned VPR backbone, not directly on features from a 3D foundation model. Our key observation is that \textit{``VGGT's DINO patch tokens become VPR-discriminative once their shared low-frequency directions are removed.''}

To the best of our knowledge, no prior work has demonstrated that DINO tokens from a 3D reconstruction model can function as reliable VPR descriptors without any fine-tuning. This training-free pipeline is crucial for SwiftVGGT, as it avoids the heavy cost of an additional VPR encoder while maintaining loop-closure accuracy comparable to fully trained models.
\vspace{-3mm}
\paragraph{Novelty.}The novelty of our Sec.~\textcolor{red}{3.3} is not merely applying normalization tricks, but lies in the following:
\vspace{2mm}
\begin{itemize}
	\item [$\bullet$] Discovering that VGGT's 3D transformer already embeds stable place-level signals, despite no VPR supervision.
	\vspace{1mm}
	\item [$\bullet$] Showing a minimal, training-free normalization pipeline unlocks these signals.
	\vspace{1mm}
	\item [$\bullet$] Replacing a dedicated VPR model entirely, reducing computation and enabling the large speedup reported in the experiments.
\end{itemize}

\begin{table*}[!t] 
	\begin{center}
		\caption{Comparison of ATE RMSE performance between the LiDAR-based method (PIN-SLAM) and RGB-based methods.}
		\vspace{-3mm}
		\resizebox{\linewidth}{!}{
			\centering
			\setlength{\tabcolsep}{6pt}
			\scriptsize
			\begin{tabular}{l||c|ccccccccccc|c}
				\toprule 
				Methods				& Modality			& 00$^\dagger$		& 01		& 02$^\dagger$		& 03		& 04		& 05$^\dagger$		& 06$^\dagger$		& 07$^\dagger$		& 08$^\dagger$		& 09$^\dagger$		& 10		& Average 		\\ \midrule \midrule
				DPV-SLAM~\cite{dpv-slam}		& RGB		& 112.80		& 11.50		& 123.53		& 2.50		& 0.81		& 57.8		& 54.86		& 18.77		& 110.49		& 76.66		& 13.65		& 53.03  \\
				DROID-SLAM~\cite{droidslam}		& RGB				& 92.10		& 344.6		& 107.61		& 2.38		& 1.00		& 118.5		& 62.47		& 21.78		& 161.60		& 72.32		& 118.70		& 100.28  \\ [1pt]
				VGGT-Long~\cite{vggtlong}		& RGB				& 9.87		& 111.06		& 37.56		& 4.89		& 3.75		& 9.09		& 7.47		& 4.02		& 62.86		& 47.48		& 25.49		& 29.41  \\ [1pt]
				SwiftVGGT (\textbf{Ours})					& RGB				& 8.17		& 102.53		& 36.49		& 8.12		& 4.88		& 11.94		& 8.88		& 5.01		& 64.68		& 44.13		& 26.18		& 29.18  \\ \midrule
				PIN-SLAM~\cite{pinslam}					& LiDAR				& \textbf{0.91}			& \textbf{3.26}		& \textbf{2.15}		& \textbf{0.43}		& \textbf{0.81}		& \textbf{0.25}		& \textbf{0.16}		& \textbf{0.25}		& \textbf{0.18}		& \textbf{1.20}		& \textbf{0.56}		& \textbf{1.07}	 \\ \bottomrule
			\end{tabular}
		}
		\label{tab:kitti_pin_slam}
	\end{center}
	\vspace{-4mm}
\end{table*}

\section{Why We Use PIN-SLAM as Pseudo GT}

A key challenge in evaluating camera tracking accuracy on the KITTI Odometry dataset~\cite{kitti} is that ground-truth poses are available only for sequences 00--10, while sequences 11--21 do not provide any pose annotations. To enable quantitative evaluation on these sequences, we adopt the poses produced by PIN-SLAM~\cite{pinslam} as pseudo ground truth.

This choice is justified by the remarkable robustness of PIN-SLAM, which leverages LiDAR measurements to estimate trajectories. Unlike purely vision-based SLAM~\cite{droidslam,dpv-slam} systems that may drift significantly in long or texture-sparse environments, PIN-SLAM benefits from accurate geometric constraints derived from high-resolution LiDAR scans. As shown in Table~\ref{tab:kitti_pin_slam}, its trajectory error remains extremely small across all kilometer-scale KITTI scenes, consistently approaching near-zero ATE values even in challenging driving scenarios. This performance is well aligned with prior observations that LiDAR-based SLAM methods can maintain long-term consistency without suffering from cumulative drift.

Because PIN-SLAM provides stable and precise poses over large-scale sequences where no official ground truth exists, its trajectories serve as a reliable surrogate for evaluating loop detection and camera tracking accuracy. Consequently, using PIN-SLAM as pseudo ground truth allows us to conduct fair and meaningful comparisons across all KITTI scenes, including those without provided annotations, while avoiding distortions that would arise from qualitative-only assessments.

\section{Depth Normalization to the Ref. Intrinsic}
\label{sec:app_depth_norm}

Our reliability-guided point sampling~Sec.~\textcolor{red}{3.2} relies on the observation that, after rescaling all depth maps to a common reference intrinsic, overlapping frames from neighboring temporal chunks produce almost identical depth values in static regions. Here we briefly derive the depth normalization in Eq.~(\textcolor{red}{1}) of the main paper.

Consider the standard pinhole camera model with focal lengths $f_x$ and $f_y$. Let $D_{\mathrm{src}}$ denote the depth map estimated by VGGT under the source intrinsic $(f_{x,\mathrm{src}}, f_{y,\mathrm{src}})$, and let $D_{\mathrm{ref}}$ be the depth that would be obtained under the reference intrinsic $(f_{x,\mathrm{ref}}, f_{y,\mathrm{ref}})$ while keeping the underlying 3D geometry fixed. Ignoring lens distortion and assuming a locally planar pixel grid, the depths are related by a simple scale factor:
\begin{equation}
	D_{\mathrm{ref}} \approx \frac{1}{2}\left(\frac{f_{x,\mathrm{ref}}}{f_{x,\mathrm{src}}} + \frac{f_{y,\mathrm{ref}}}{f_{y,\mathrm{src}}}\right) D_{\mathrm{src}}.
\end{equation}
We adopt this symmetric scaling to account for potential anisotropy in $f_x$ and $f_y$. Empirically, this normalization makes the overlapping regions between neighboring temporal chunks highly consistent in depth, which is crucial for our subsequent reliability-guided sampling strategy.

\section{Reliability-Guided Point Sampling}

After normalizing depths to the reference intrinsic, most pixels in the overlapping regions of neighboring temporal chunks already exhibit small depth discrepancies, while large deviations are concentrated near object boundaries and occlusion edges. Our reliability-guided point sampling exploits this structure by retaining only pixels whose depth difference and confidence satisfy
\begin{equation}
	\begin{aligned}
		& \big| D_{\mathrm{ref},t} - D_{\mathrm{ref},t+1} \big| < \lambda_D, \\
		& \gamma_t > \lambda_{\gamma} \mu_{\gamma_t}, \\
		& \gamma_{t+1} > \lambda_{\gamma} \mu_{\gamma_{t+1}}.
	\end{aligned}
\end{equation}

In practice, our sampling strategy does not aggressively sparsify the overlap region. Instead, it selects pixels whose depths, after being normalized to the reference intrinsic, are mutually consistent between neighboring chunks and have sufficiently high confidence according to VGGT's depth-confidence head. This effectively removes depth-discontinuous boundaries and low-confidence regions, while retaining the geometrically stable portions of the overlap. As a result, the correspondences fed to the Umeyama solver are both reliable and well aligned, which explains the substantial runtime reduction reported in Tab.~\textcolor{red}{5} of the main paper and the improved ATE compared to using all points.

\section{Additional Ablation Studies.}
We conduct further ablation experiments to supplement the analyses presented in the main paper.

\paragraph{Loop detection.}\cref{tab:additional_ablation_loop_detection} reports an ablation study on the loop detection pipeline using the 12 KITTI sequences containing loops. Removing loop detection entirely eliminates the need for loop-centric chunk inference, reducing runtime by approximately 16 seconds, but results in a substantial degradation of ATE (over 16~m). A naive baseline that directly averages VGGT's DINO patch tokens followed by $\ell_2$ normalization achieves better accuracy than the no-loop variant, but still performs worse than our full pipeline due to incorrect loop associations. Adding signed power normalization improves performance by producing more balanced and discriminative descriptors. Finally, incorporating PCA whitening yields the best overall ATE, confirming that both power normalization and whitening are necessary to transform VGGT patch tokens into reliable loop descriptors without requiring a dedicated VPR encoder.

\paragraph{Reliability-guided point sampling.}\cref{tab:additional_ablation_rgps} presents a more detailed ablation of our reliability-guided point sampling strategy. Using all overlap pixels leads to suboptimal alignment accuracy and slightly higher runtime due to the large number of point correspondences. Applying only the depth difference threshold already produces the best accuracy, demonstrating that depth alignment alone is a strong reliability indicator. We additionally retain depth confidence filtering because, despite a marginal decrease in average ATE, it yields more stable performance across all scenes and generalizes better to the Waymo Open~\cite{waymo} and Virtual KITTI datasets~\cite{vkitti}. This trade-off makes it the preferred configuration in our final model.

\begin{table}[!t] 
	\begin{center}
		\caption{Additional ablation study on loop detection methods.}
		\vspace{-3mm}
		\resizebox{0.9\columnwidth}{!}{
			\centering
			\setlength{\tabcolsep}{3pt}
			\scriptsize
			\begin{tabular}{l|cc}
				\toprule
				Method								& Average ATE (m)			& Elapsed Time (s) \\ \midrule \midrule
				w/o Loop Detection			& 44.01									& \textbf{126.35} \\ [1pt]
				w/ Token Avg. + $\ell_2$ Norm. & 34.39						& 142.98	\\ [1pt]
				w/ Signed Power					& 27.77									& 143.13 \\ \midrule
				Full (Ours)								& \textbf{27.33}									& 142.67 \\ \bottomrule
			\end{tabular}
		}
		\label{tab:additional_ablation_loop_detection}
	\end{center}
	\vspace{-6mm}
\end{table}

\section{Additional Results}

\paragraph{Virtual KITTI dataset.}We evaluate our method on the Virtual KITTI dataset~\cite{vkitti}, and the quantitative results are summarized in~\cref{tab:comparison_vkitti}. Although DROID-SLAM~\cite{droidslam} achieves the best average ATE, it completely fails to track camera poses in the \texttt{0006\_rain} sequence. CUT3R~\cite{cut3r} and FastVGGT~\cite{fastvggt} show poor tracking results, exceeding $10$\,m ATE in most sequences, and FastVGGT additionally runs out of GPU memory on \texttt{0020}. In contrast, both VGGT-Long~\cite{vggtlong} and our method maintain consistently strong performance across all scenes. Notably, SwiftVGGT achieves slightly better accuracy while being significantly faster, demonstrating its robustness even in short, loop-free sequences.

\begin{table}[!t] 
	\begin{center}
		\caption{Additional ablation study on loop detection methods.}
		\vspace{-3mm}
		\resizebox{0.7\columnwidth}{!}{
			\centering
			\setlength{\tabcolsep}{8pt}
			\scriptsize
			\begin{tabular}{l|c}
				\toprule
				Method												& Average ATE (m)			 \\ \midrule \midrule
				w/ IRLS												& 29.23															\\ \midrule
				w/ Whole Points								& 32.95																	\\ [1pt]
				w/ Depth Diff. Only							& \textbf{28.42}																\\ [1pt]
				w/ Depth Conf. Only						& 30.23																	\\ \midrule
				Full (Ours)										& 29.18								 \\ \bottomrule
			\end{tabular}
		}
		\label{tab:additional_ablation_rgps}
	\end{center}
	\vspace{-6mm}
\end{table}

\paragraph{Waymo Open dataset.}Following the evaluation protocol described in the main paper, we report 3D reconstruction metrics in~\cref{tab:waymo_point}. Since the ground-truth point clouds are generated directly from Waymo~\cite{waymo} LiDAR measurements and camera parameters, their density and coverage differ from the reconstructed outputs. As a result, completeness scores are generally low across all methods. MASt3R-SLAM and CUT3R yield relatively large errors, whereas FastVGGT, VGGT-Long, and SwiftVGGT achieve comparably strong results. Considering the point cloud metrics in this table together with the camera tracking performance and runtime comparisons in Tab.~\textcolor{red}{3}, SwiftVGGT offers the most favorable balance of accuracy and efficiency.

\paragraph{KITTI Scene 11--20.}Using pseudo ground-truth trajectories generated by PIN-SLAM~\cite{pinslam}, we evaluate scenes 11--20 of the KITTI dataset~\cite{kitti}, covering both loop and non-loop sequences. The results are presented in~\cref{tab:comparison_additional_kitti}. Regardless of whether loops are present, SwiftVGGT consistently outperforms prior SLAM-based and VGGT-based methods. These results highlight that our loop detection mechanism remains effective when loops exist, but does not degrade performance when they do not, further demonstrating the general applicability of our approach.

\begin{table*}[!t] 
	\begin{center}
		\caption{Performance comparison on the Waymo Open dataset~\cite{waymo}. Acc. and Comp. denote accuracy and completeness, respectively; CD stands for Chamfer Distance.}
		\vspace{-3mm}
		\resizebox{\linewidth}{!}{
			\centering
			\setlength{\tabcolsep}{7pt}
			\scriptsize
			\begin{tabular}{l||ccc|ccc|ccc|ccc|ccc}
				\toprule
				\multirow{2}{*}{Methods}	& \multicolumn{3}{c|}{163453191}	& \multicolumn{3}{c|}{183829460}	& \multicolumn{3}{c|}{315615587}	& \multicolumn{3}{c|}{346181117}	& \multicolumn{3}{c}{371159869} \\ [1pt]
				& Acc.	& Comp.	& CD				& Acc.	& Comp.	& CD					& Acc.	& Comp.	& CD				& Acc.	& Comp.	& CD				& Acc.	& Comp.	& CD	\\ \midrule \midrule
				%				DROID-SLAM~\cite{droidslam}			& 	& 	& 											& 	& 	&														& 	& 	&													& 	& 	&												& 	& 	&\\ [1pt]
				MASt3R-SLAM~\cite{mast3r-slam}	& 0.475	& 35.440 & 17.960 		& 0.560 & 17.790 & 9.180						& 1.003 & 12.210 & 6.608				& 1.554 & 9.980 & 5.767					& \cellcolor{best!25}0.810 & 32.001 & 16.405 \\ [1pt]
				CUT3R~\cite{cut3r}								& 3.015	& 8.071 & 5.543 			& 0.485	& 11.700 & 6.093						& 1.667	& 8.556 & 5.117					& 3.216 & 11.450 & 7.333				& 2.801 & 13.062 & 7.932\\ [1pt]
				FastVGGT~\cite{fastvggt}					& \cellcolor{best!25}0.349	& \cellcolor{best!25}3.157	& \cellcolor{best!25}1.753				& \cellcolor{second!35}0.305	& \cellcolor{best!25}8.058	& \cellcolor{best!25}4.182							& 0.738	& \cellcolor{best!25}4.845	& \cellcolor{second!35}2.792					& 0.882	& \cellcolor{best!25}4.601	& \cellcolor{second!35}2.741					& \cellcolor{second!35}0.937	& 8.170	& 4.554\\ [1pt]
				VGGT-Long~\cite{vggtlong}				& \cellcolor{second!35}0.401	& 4.319	& 2.360				& \cellcolor{best!25}0.200	& 9.489	& \cellcolor{second!35}4.844							& \cellcolor{best!25}0.222	& 5.409	& 2.816					& \cellcolor{best!25}0.326	& \cellcolor{second!35}4.932	& \cellcolor{best!25}2.629					& 1.113	& \cellcolor{second!35}7.880	& \cellcolor{second!35}4.497\\ \midrule
				Ours														& 0.492	& \cellcolor{second!35}4.171	& \cellcolor{second!35}2.331				& 0.595	& \cellcolor{second!35}9.150	& 4.872								& \cellcolor{second!25}0.329	& \cellcolor{second!35}5.181	& \cellcolor{best!25}2.755					& \cellcolor{second!35}0.451	& 5.552	& 3.001					& 1.234	& \cellcolor{best!25}7.215	& \cellcolor{best!25}4.224\\ \midrule \midrule
				
				\multirow{2}{*}{Methods}	& \multicolumn{3}{c|}{405841035}	& \multicolumn{3}{c|}{460417311}	& \multicolumn{3}{c|}{520018670}	& \multicolumn{3}{c|}{610454533}	& \multicolumn{3}{c}{Average} \\ [1pt]
				& Acc.	& Comp.	& CD				& Acc.	& Comp.	& CD					& Acc.	& Comp.	& CD				& Acc.	& Comp.	& CD				& Acc.	& Comp.	& CD		\\ \midrule \midrule
				%				DROID-SLAM~\cite{droidslam}			& 	& 	& 											& 	& 	&														& 	& 	&													& 	& 	&												& 	& 	&\\ [1pt]
				MASt3R-SLAM~\cite{mast3r-slam}	& 0.284 & 11.630 & 5.959 		& 1.109 & 5.080 & 3.095							& 1.015	& 12.810 & 6.913						& 0.840	& 5.889	&	3.365			& 0.850	& 15.870 & 8.361\\ [1pt]
				CUT3R~\cite{cut3r}								& 2.008	& 7.947	& 4.977				& 1.335	& 6.501	& 3.918							& 1.567	& \cellcolor{best!25}9.337	& 5.452						& 1.003	& 35.220 & 18.110				& 1.900	& 12.428 & 7.164\\ [1pt]
				FastVGGT~\cite{fastvggt}					& 0.243	& \cellcolor{best!25}6.668	& \cellcolor{best!25}3.455				& 0.707	& \cellcolor{best!25}2.827	& \cellcolor{best!25}1.767							& 0.624	& 9.748	& 5.186						& 0.774	& \cellcolor{best!25}5.204	& \cellcolor{second!35}2.989					& 0.618	& \cellcolor{best!25}5.920	& \cellcolor{best!25}3.269\\ [1pt]
				VGGT-Long~\cite{vggtlong}				& \cellcolor{second!35}0.185	& \cellcolor{second!35}7.434	& \cellcolor{second!35}3.810				& \cellcolor{best!25}0.197	& 3.604	& 1.900							& \cellcolor{best!25}0.251	& 9.558	& \cellcolor{best!25}4.904						& \cellcolor{second!35}0.487	& \cellcolor{second!35}5.287	& \cellcolor{best!25}2.887					& \cellcolor{best!25}0.376	& 6.435	&\cellcolor{second!35}3.405\\ \midrule
				Ours														& \cellcolor{best!25}0.151	& 7.570	& 3.861					& \cellcolor{second!35}0.547	& \cellcolor{second!35}3.172	& \cellcolor{second!35}1.859							& \cellcolor{second!35}0.333	& \cellcolor{second!35}9.488	& \cellcolor{second!35}4.910						& \cellcolor{best!25}0.440	& 5.537	& \cellcolor{second!35}2.989				& \cellcolor{second!35}0.508	& \cellcolor{second!35}6.337	& 3.422\\ \bottomrule
			\end{tabular}
		}
		\label{tab:waymo_point}
	\end{center}
	\vspace{-5mm}
\end{table*}

\begin{table*}[!t] 
	\begin{center}
		\caption{Performance comparison on the Virtual KITTI dataset~\cite{vkitti}. The \colorbox{best!25}{orange} and \colorbox{second!35}{yellow} cells respectively indicate the highest and second-highest value.}
		\vspace{-3mm}
		\resizebox{\linewidth}{!}{
			\centering
			\setlength{\tabcolsep}{4pt}
			\scriptsize
			\begin{tabular}{l||cccccc|cccccc|cccccc}
				\toprule
				\multirow{2}{*}{Methods}				& \multicolumn{6}{c|}{Scene 0001}															& \multicolumn{6}{c|}{Scene 0002}														& \multicolumn{6}{c}{Scene 0006}	 \\ [1pt]
																			& Clone	& Fog	& Morning	& Overcast	& Rain	& Sunset				& Clone	& Fog	& Morning	& Overcast	& Rain	& Sunset				& Clone	& Fog	& Morning	& Overcast	& Rain	& Sunset	\\ \midrule \midrule
				DROID-SLAM~\cite{droidslam}		& 1.027 & 1.868 & \cellcolor{second!35}0.989 & \cellcolor{second!35}1.015 & \cellcolor{second!35}0.776 & \cellcolor{second!35}1.145 								& \cellcolor{best!25}0.098 & \cellcolor{best!25}0.040 & \cellcolor{best!25}0.049 & \cellcolor{best!25}0.048 & \cellcolor{best!25}0.036 & \cellcolor{best!25}0.113 						& \cellcolor{best!25}0.063 & \cellcolor{best!25}0.024 & \cellcolor{best!25}0.030 & \cellcolor{best!25}0.051 & TL & \cellcolor{best!25}0.0197\\ [1pt]
				CUT3R~\cite{cut3r}						& 43.304 & 62.191 & 50.608 & 38.744 & 51.548 & 43.785 					& 23.771 & 9.948 & 28.415 & 24.644 & 7.963 & 25.973 				& 0.836 & \cellcolor{second!35}0.408 & 0.599 & 0.720 & 1.059 & 1.013\\ [1pt]
				FastVGGT~\cite{fastvggt}			& 57.097 & 58.584 & 59.552 & 56.839 & 58.383 & 57.999 					& 2.958 & 2.997 & 2.946 & 2.965 & 2.974 & 2.955 						& 3.731 & 3.812 & 3.731 & 3.742 & 3.792 & 3.732\\ [1pt]
				VGGT-Long~\cite{vggtlong}		& \cellcolor{best!25}0.838 & \cellcolor{second!35}1.288 & \cellcolor{best!25}0.791 & \cellcolor{best!25}0.755 & 1.474 & \cellcolor{best!25}0.894 								& \cellcolor{second!35}0.702 & 0.674 & \cellcolor{second!35}0.734 & 0.662 & \cellcolor{second!35}0.673 & \cellcolor{second!35}0.656 						& 0.435 & 0.549 & 0.430 & 0.439 & \cellcolor{second!35}0.534 & 0.444\\ \midrule
				Ours												& \cellcolor{second!35}0.888 & \cellcolor{best!25}1.170 & 1.668 & 0.797 & \cellcolor{second!35}1.431 & 1.792 & 0.725 & \cellcolor{second!35}0.637 & 0.747 & \cellcolor{second!35}0.630 & 0.689 & 0.687 & \cellcolor{second!35}0.398 & 0.433 & \cellcolor{second!35}0.396 & \cellcolor{second!35}0.398 & \cellcolor{best!25}0.461 & \cellcolor{second!35}0.409\\  \midrule\midrule
				
				\multirow{2}{*}{Methods}				& \multicolumn{6}{c|}{Scene 0018}															& \multicolumn{6}{c|}{Scene 0020}														& \multicolumn{6}{c}{Average}	 \\ [1pt]
																			& Clone	& Fog	& Morning	& Overcast	& Rain	& Sunset				& Clone	& Fog	& Morning	& Overcast	& Rain	& Sunset				& 0001	& 0002	& 0006	& 0018	& 0020	& Average	\\ \midrule \midrule
				DROID-SLAM~\cite{droidslam}		& 2.478 & 2.032 & 1.894 & 2.332 & 2.549 & \cellcolor{second!35}1.943 							& \cellcolor{best!25}3.592 & \cellcolor{best!25}5.079 & \cellcolor{second!35}3.733 & \cellcolor{second!35}3.852 & \cellcolor{best!25}3.780 & \cellcolor{second!35}4.907 						& \cellcolor{second!35}1.137 & \cellcolor{best!25}0.064 & \cellcolor{best!25}0.038 & 2.205 & \cellcolor{best!25}4.157 & \cellcolor{best!25}1.571\\ [1pt]
				CUT3R~\cite{cut3r}						& 19.440 & 8.628 & 6.720 & 20.212 & 16.777 & 31.119 						& 129.498 & 76.962 & 117.948 & 114.512 & 66.700 & 116.529 		& 48.363 & 20.119 & 0.772 & 17.149 & 103.692 & 38.019\\ [1pt]
				FastVGGT~\cite{fastvggt}			& 4.781 & 4.795 & 4.662 & 5.449 & 4.570 & 4.707 								& OOM & OOM & OOM & OOM & OOM & OOM 								& 58.076 & 2.966 & 3.762 & 4.827 & OOM & 17.406\\ [1pt]
				VGGT-Long~\cite{vggtlong}		& \cellcolor{second!35}1.778 & \cellcolor{second!35}0.796 & \cellcolor{best!25}1.324 & \cellcolor{best!25}1.425 & \cellcolor{second!35}1.900 & 1.998 								& 13.094 & 14.762 & 6.649 & \cellcolor{best!25}3.460 & 5.191 & \cellcolor{best!25}3.749 							& \cellcolor{best!25}1.006 & \cellcolor{second!35}0.684 & 0.472 & 1.537 & 7.817 & 2.303\\ \midrule
				Ours												& \cellcolor{best!25}1.747 & \cellcolor{best!25}0.787 & \cellcolor{second!35}1.516 & \cellcolor{second!35}1.653 & \cellcolor{best!25}1.572 & \cellcolor{second!35}1.780 & \cellcolor{second!35}9.599 & \cellcolor{second!35}14.087 & \cellcolor{best!25}3.141 & 5.969 & \cellcolor{second!35}5.031 & 6.339 & 1.291 & 0.686 & \cellcolor{second!35}0.416 & \cellcolor{best!25}1.509 & \cellcolor{second!35}7.361 & \cellcolor{second!35}2.253\\ \bottomrule
			\end{tabular}
		}
		\label{tab:comparison_vkitti}
	\end{center}
	\vspace{-5mm}
\end{table*}

\begin{table*}[!t] 
	\begin{center}
		\caption{Performance comparison on the KITTI Odometry dataset~\cite{kitti} Scene 11--20. The $^{\dagger}$ denotes sequences containing loops, and \textit{Dense}$^\diamondsuit$ indicates semi-dense 3D reconstruction.}
		\vspace{-3mm}
		\resizebox{\linewidth}{!}{
			\centering
			\setlength{\tabcolsep}{7pt}
			\scriptsize
			\begin{tabular}{l||c|c|cccccccccc|c}
				\toprule 
				Methods												& Calib.		& Recon.												& 11		& 12		& 13$^\dagger$		& 14		& 15$^\dagger$		& 16$^\dagger$		& 17		& 18$^\dagger$		& 19$^\dagger$		& 20			& Average 	\\ \midrule \midrule
				DPV-SLAM~\cite{dpv-slam}		& \ding{51}		& \textit{Sparse}									& \cellcolor{second!35}26.00 & 57.06 & 53.73 		& 9.72 		& 51.20 		& 34.15 		& 57.29 		& 17.96 			& 229.36 		& \cellcolor{second!35}8.73 				& 54.52\\ \midrule
				DROID-SLAM~\cite{droidslam}		& \ding{51}		& \textit{Dense}$^\diamondsuit$		& \cellcolor{best!25}17.18 & \cellcolor{best!25}9.61 & 48.40 			& \cellcolor{second!35}8.93 		& 39.97 		& 30.41 		& \cellcolor{best!25}0.86 			& 11.58 				& 182.82 				& \cellcolor{best!25}1.04 					& 35.08\\ [1pt]
				VGGT-Long~\cite{vggtlong}		& \ding{55}		& \textit{Dense}									& 38.45 & \cellcolor{second!35}25.93 & \cellcolor{second!35}7.76 			& 12.98 		& \cellcolor{second!35}6.33 		& \cellcolor{second!35}18.71 		& 22.22 			& \cellcolor{best!25}6.84 				& \cellcolor{second!35}118.10 				& 10.13 					& \cellcolor{second!35}26.75\\ \midrule
				Ours													& \ding{55}		& \textit{Dense}									& 38.77 & 36.42 & \cellcolor{best!25}6.72 		& \cellcolor{best!25}8.38 		& \cellcolor{best!25}6.30 		& \cellcolor{best!25}16.27 		& \cellcolor{second!35}17.59 		& \cellcolor{second!35}7.60 				& \cellcolor{best!25}111.73 				& 16.73 				& \cellcolor{best!25}26.65\\ \bottomrule
			\end{tabular}
		}
		\label{tab:comparison_additional_kitti}
	\end{center}
	\vspace{-4mm}
\end{table*}

\begin{figure*}[t]
	\centering
	\includegraphics[width=0.95\textwidth]{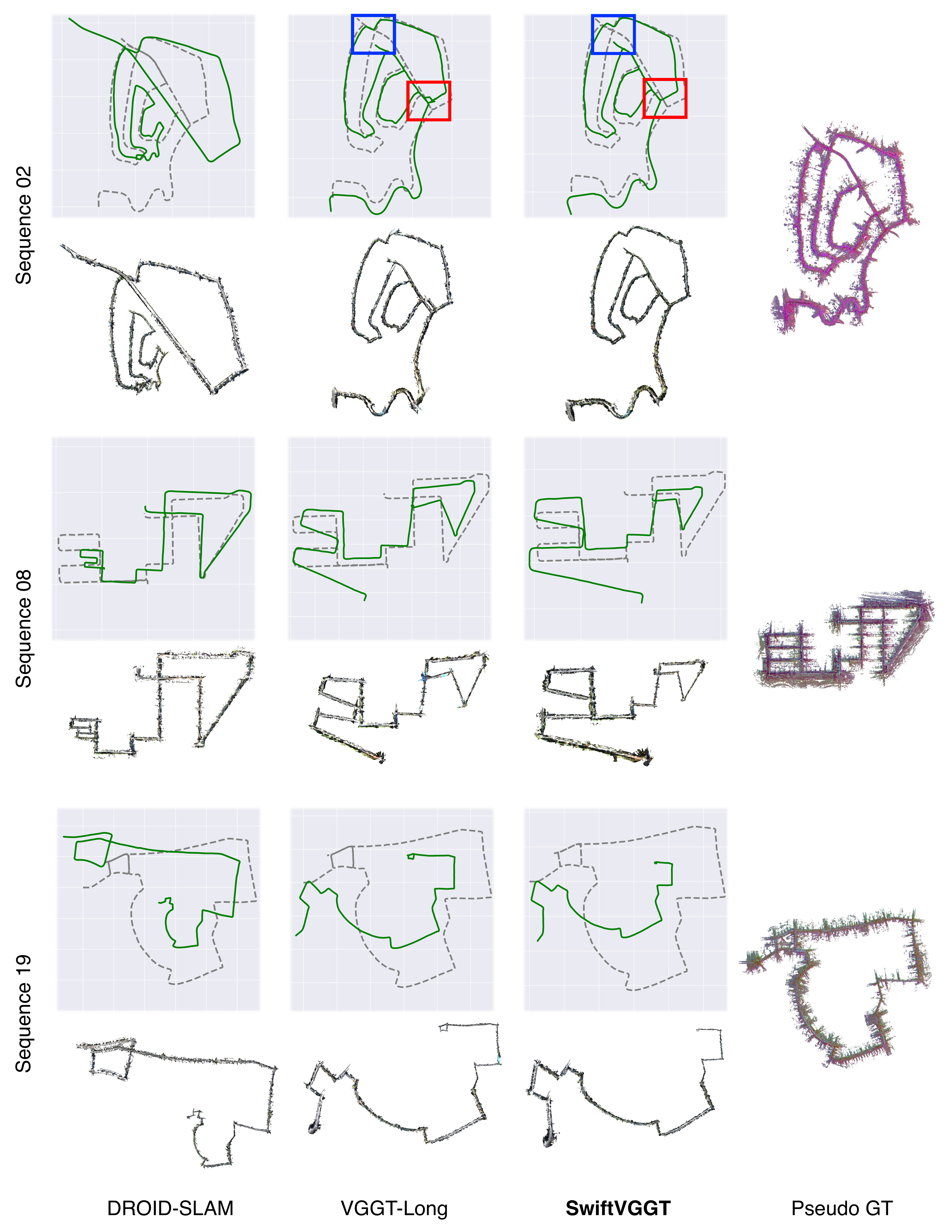}
	\vspace{-3mm}
	\caption{Failure cases of KITTI scenes. The pseudo ground-truth point cloud is obtained by combining the LiDAR point cloud with camera poses either provided by the dataset or estimated by PIN-SLAM~\cite{pinslam}. The \textcolor{grayline}{\textbf{gray}} dashed line represents the ground-truth trajectory, while the \textcolor{greenline}{\textbf{green}} solid line denotes the estimated trajectory.}
	\label{fig:failure_cases}
	\vspace{-3mm}
\end{figure*}

\clearpage
\clearpage

\begin{figure}[t]
	\centering
	\includegraphics[width=\linewidth]{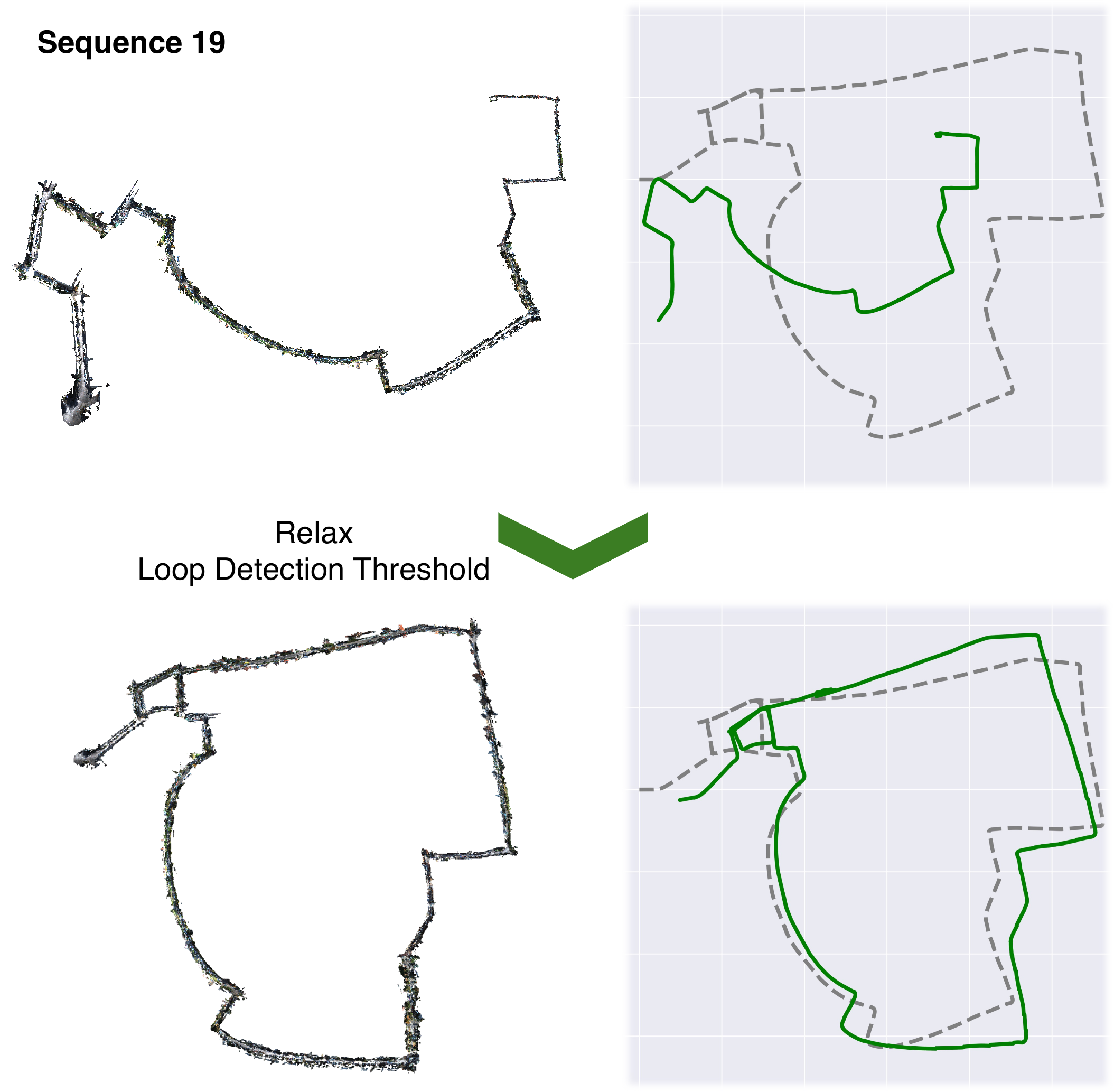}
	\vspace{-3mm}
	\caption{Point cloud visualization of KITTI dataset~\cite{kitti}.}
	\label{fig:threshold}
	\vspace{-3mm}
\end{figure}

\section{Failure Case Analysis}

In this section, we analyze failure cases observed in several KITTI~\cite{kitti} sequences. As illustrated in~\cref{fig:failure_cases}, SwiftVGGT, DROID-SLAM~\cite{droidslam}, and VGGT-Long~\cite{vggtlong} all exhibit degraded performance on sequences 02, 08, and 19, both quantitatively and qualitatively. A common characteristic of these sequences is that loop closure is not successfully detected. For instance, in KITTI~02, DROID-SLAM completely diverges, while VGGT-Long incorrectly detects non-existent loops (\textcolor{red}{\textbf{red}} box in~\cref{fig:failure_cases}) and simultaneously misses true loops (\textcolor{blue}{\textbf{blue}} box). In contrast, our method avoids hallucinated loops but still fails to identify the missing true loops. Similarly, in sequences 08 and 19, all three methods consistently fail to detect valid loop closures. Interestingly, as shown in~\cref{fig:threshold}, relaxing the loop detection threshold enables SwiftVGGT to correctly detect the loop in sequence~19, resulting in noticeably improved reconstruction quality. However, this adjustment does not reliably solve all failure cases across all sequences. Nonetheless, these results suggest that VGGT’s DINO transformer features can outperform classical VPR encoders when appropriately processed, even without training or fine-tuning. A promising direction for future work is to incorporate feature-level or correspondence-level loop detection strategies (\textit{e.g.}, point tracking or feature matching) to further improve robustness under challenging loop conditions.

Finally, beyond loop detection failures, some sequences still exhibit relatively high ATE (greater than 10m even when loops are correctly found). As discussed in Sec.~\textcolor{red}{5} of the main paper, this limitation largely stems from the absence of bundle adjustment, meaning accumulated drift remains uncorrected. Integrating a lightweight or learned BA module into SwiftVGGT may address this remaining challenge and enable more precise large-scale reconstruction.

\section{Point Cloud Visualization}

\begin{figure*}[t]
	\centering
	\includegraphics[width=0.95\textwidth]{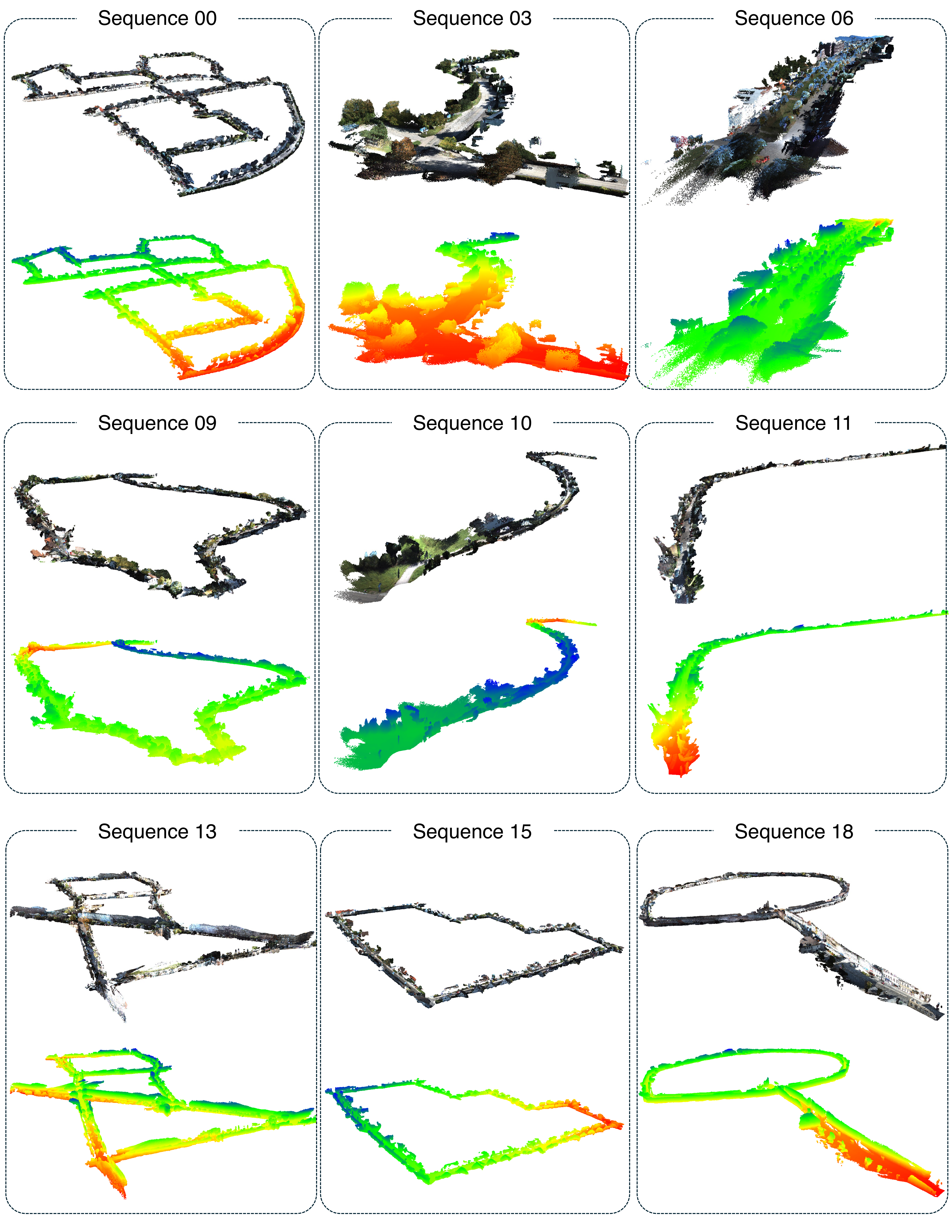}
	\vspace{-3mm}
	\caption{Point cloud visualization of KITTI dataset~\cite{kitti}.}
	\label{fig:qualitative_kitti}
	\vspace{-3mm}
\end{figure*}

\begin{figure*}[t]
	\centering
	\includegraphics[width=0.95\textwidth]{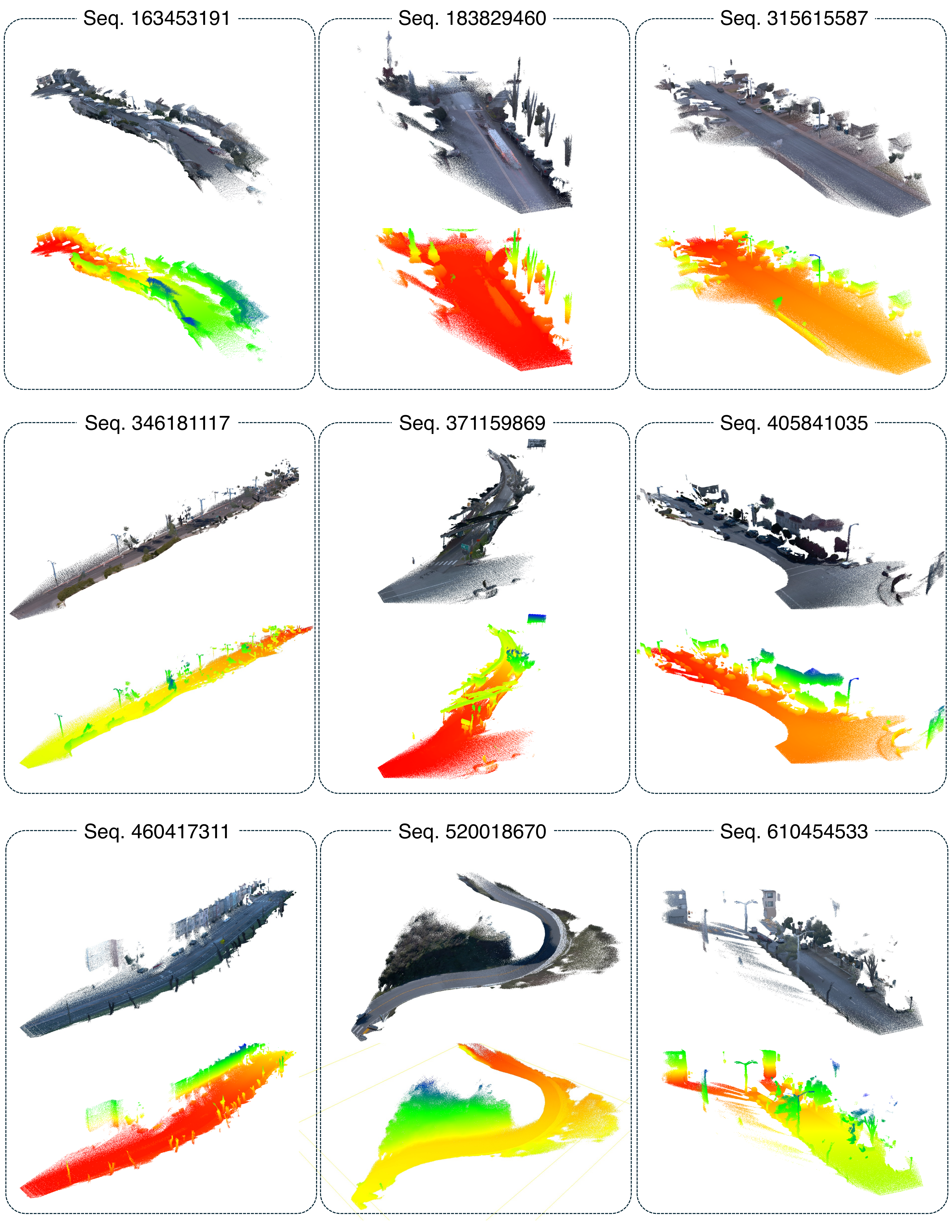}
	\vspace{-3mm}
	\caption{Point cloud visualization of Waymo Open dataset~\cite{waymo}.}
	\label{fig:qualitative_waymo}
	\vspace{-3mm}
\end{figure*}

\begin{figure*}[t]
	\centering
	\includegraphics[width=0.95\textwidth]{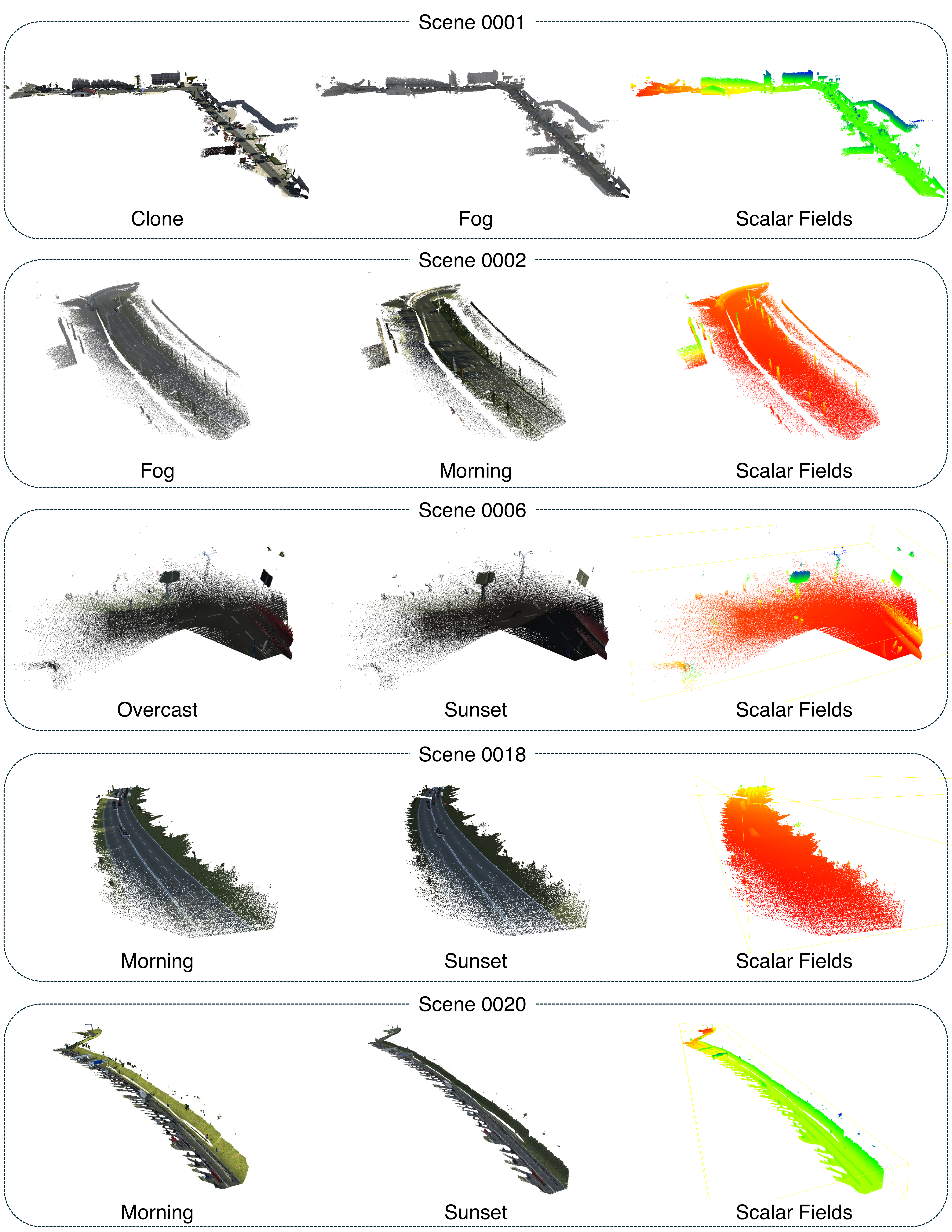}
	\vspace{-3mm}
	\caption{Point cloud visualization of Virtual KITTI dataset~\cite{vkitti}.}
	\label{fig:qualitative_vkitti}
	\vspace{-3mm}
\end{figure*}

We provide qualitative dense reconstruction results on the KITTI~\cite{kitti}, Waymo Open~\cite{waymo}, and Virtual KITTI~\cite{vkitti} datasets in Figures~\cref{fig:qualitative_kitti}, \cref{fig:qualitative_waymo}, and \cref{fig:qualitative_vkitti}. The reconstructed point clouds are visualized with a depth-based color map, where the color corresponds to the $z$-axis value of each point. These results demonstrate that, in addition to achieving fast inference and state-of-the-art tracking accuracy, SwiftVGGT is capable of producing high-quality dense 3D reconstructions across diverse large-scale driving scenarios.

\clearpage
{
    \small
    \bibliographystyle{ieeenat_fullname}
    \bibliography{main}
}

% WARNING: do not forget to delete the supplementary pages from your submission 
% \input{sec/X_suppl}

\end{document}